%% file: main.tex
\documentclass{article}

\PassOptionsToPackage{numbers, compress}{natbib}

\usepackage[main, final]{neurips_2026}

\usepackage[utf8]{inputenc} %
\usepackage[T1]{fontenc}    %
\usepackage{hyperref}       %
\usepackage{url}            %
\usepackage{booktabs}       %
\usepackage{amsfonts}       %
\usepackage{nicefrac}       %
\usepackage{microtype}      %
\usepackage{xcolor}         %

\usepackage{algorithm}
\usepackage{algorithmic}
\usepackage{multirow}

\usepackage{amsmath}
\usepackage{amssymb}
\usepackage{mathtools}
\usepackage{amsthm}
\usepackage{bbm}
\usepackage{graphicx}
\usepackage[capitalize,noabbrev]{cleveref}

\usepackage{wrapfig}
\usepackage{subcaption}

\DeclareMathOperator*{\argmax}{arg\,max}

\newcommand{\beq}{ \begin{equation} }
\newcommand{\eeq}{ \end{equation} }

\newcommand{\codeurl}{https://github.com/NohyunKi/InfCLIP}

\title{\textsc{InfCLIP}: Unified Data Valuation for CLIP Pretraining via Influence-Inspired Scoring}

\author{%
  Nohyun Ki\\
  School of Electrical Engineering\\
  KAIST\\
  \texttt{kinohyun@kaist.ac.kr}
   \And
  Hye Won Chung\thanks{Corresponding author.}\\
  School of Electrical Engineering\\
  KAIST\\
  \texttt{hwchung@kaist.ac.kr}
}

\begin{document}

\maketitle
\begin{abstract}
    \input{sections/0_abstract}
\end{abstract}

\section{Introduction}
\input{sections/1_Intro}

\section{Related work}
\input{sections/2_Related}

\section{Method}
\input{sections/3_Method}

\section{Experiment}
\input{sections/4_Experiment}

\section{Conclusion}

\input{sections/5_Conclusion}

\begin{ack}
This work was supported by the National Research Foundation of Korea (NRF) grant funded by the Korea government (MSIT) (No. RS-2024-00408003 and RS-2025-00516153) and the Institute for Information \& communications Technology Planning \& Evaluation (IITP) grant funded by the Korea government (MSIT) (No. RS-2024-00444862 and RS-2026-25522672).
\end{ack}

\bibliographystyle{splncs04}
\bibliography{main}

\newpage
\appendix
\input{appendix/limitations}
\input{appendix/infclip_score}

\input{appendix/ablation_calculation}

\input{appendix/robustness}

\input{appendix/experiment_details}
\input{appendix/real_examples}

\input{appendix/full_results}

\end{document}

%% file: sections/0_abstract.tex
Effective data valuation is essential for machine learning, especially for CLIP pretraining, where models are trained on massive and inherently noisy image-text datasets.
Existing CLIP data selection methods often combine heuristic signals such as alignment, target similarity, and diversity, but lack a unified principle for estimating sample contribution.
This can lead to inconsistent performance across data regimes and obscures how each sample contributes to downstream performance.
In this work, we propose \textsc{InfCLIP}, a principled influence-inspired approach for data selection in CLIP pretraining.
\textsc{InfCLIP} estimates the contribution of each training sample to a target task by leveraging feature representations and temperature-rescaled softmax distributions, enabling efficient computation without retraining or access to model internals.
This formulation provides a unified view of alignment, informativeness, uniqueness, and target relevance within a single framework.
Empirically, \textsc{InfCLIP} consistently outperforms existing data selection methods across a wide range of benchmarks, including zero-shot evaluation on ImageNet and distribution-shifted datasets.
We also show that Self-\textsc{InfCLIP} enables efficient training-set analysis, including noise detection and memorization-aware data assessment, without additional retraining.

%% file: sections/1_Intro.tex
Data valuation aims to identify useful training examples from large datasets while filtering out noisy or unhelpful data \cite{shapley}.
This problem is particularly important for CLIP-style vision-language pretraining, where models are trained on web-scale image-text pairs that are massive, diverse, and inherently noisy \cite{CLIP, DataComp}.
Consequently, numerous data selection and filtering methods have been proposed to retain high-quality image-text pairs within the CLIP pretraining framework \cite{SCLIP, HYPE, CLIPCOV, DFN}.

\begin{figure*}[t]
    \centering

    \begin{subfigure}[b]{0.64\linewidth}
        \centering
        \includegraphics[width=\linewidth]{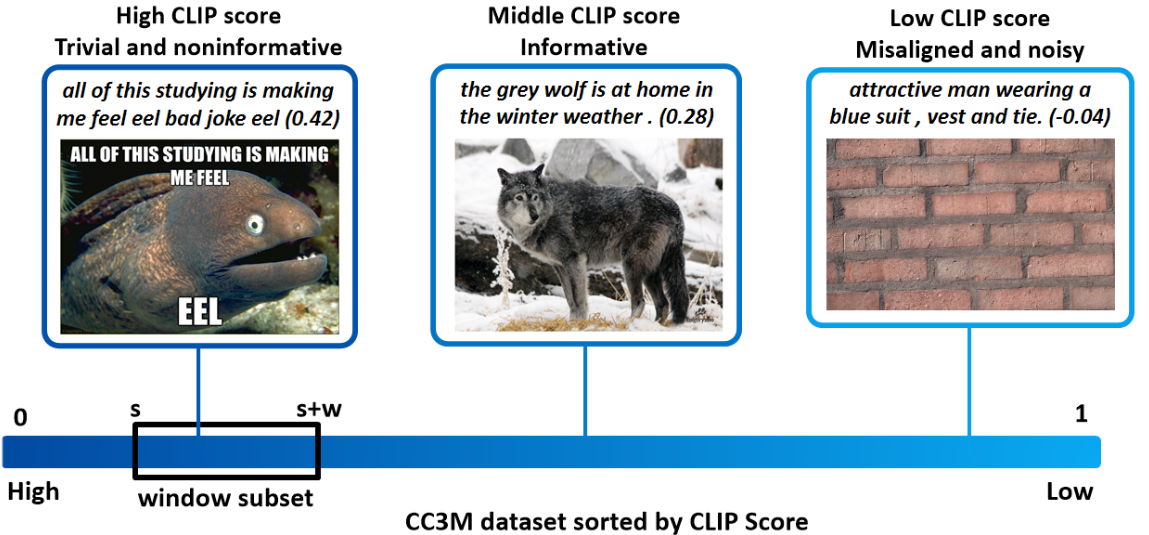}
        \caption{Qualitative examples}
        \label{fig:real_example}
    \end{subfigure}
    \hfill
    \begin{subfigure}[b]{0.28\linewidth}
        \centering
        \includegraphics[width=\linewidth]{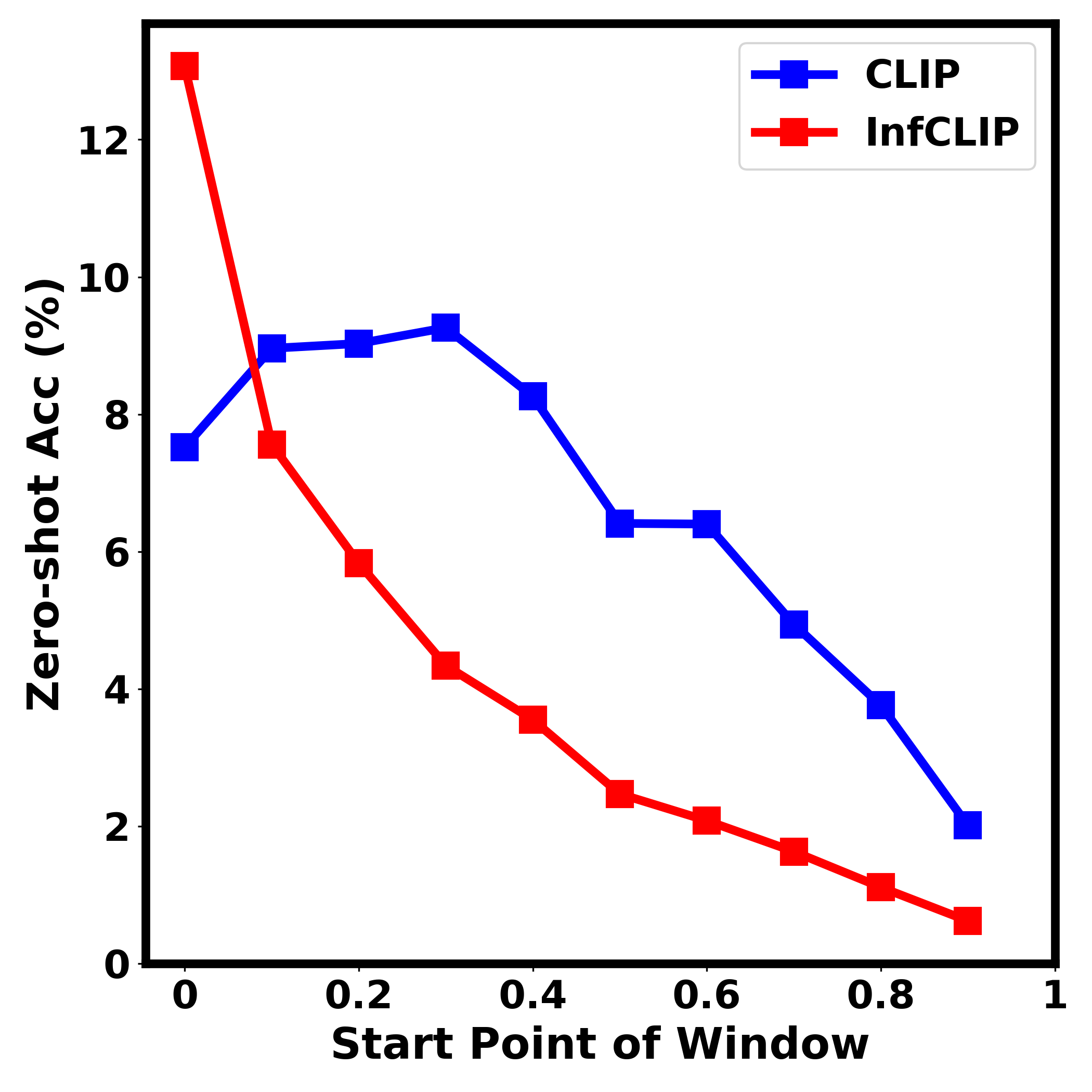}
        \caption{Window subset training}
        \label{fig:window}
    \end{subfigure}

    \caption{Alignment alone is insufficient for data valuation.
    (a) Low-CLIP-score samples are often noisy or mismatched, while very high-score samples can be trivial or redundant.
    (b) Window training experiment: after sorting data by CLIP score, each model is trained on a contiguous 10\%-sized percentile window $[s, s+10\%]$, where the x-axis denotes the starting percentile $s$.
    Mid-score windows outperform the highest-score window, whereas \textsc{InfCLIP}-based selection yields monotonic gains.}
    \label{fig:Fig1}
    \vspace{-0.2in}
\end{figure*}

Most existing data scoring methods for CLIP pretraining primarily rely on \emph{alignment}, often referred to as the \emph{CLIP score}, measured by the cosine similarity between image and text features from a pretrained teacher model.
A high alignment score indicates that the image and caption are semantically consistent, while a low score often corresponds to a noisy or mismatched image-text pair.
Thus, alignment is effective for filtering clearly corrupted samples, but it does not necessarily measure how informative a sample is for learning transferable representations.
As shown in Figure~\ref{fig:Fig1}, low-score samples are often noisy, but very high-score samples can be visually or linguistically trivial, while mid-score samples often provide richer training signals.
We further verify this through a window training experiment following \cite{BWS}: after sorting data by CLIP score, we train CLIP models on contiguous 10\%-sized percentile windows.
The highest-score window does not yield the best performance, demonstrating that alignment alone is insufficient for data valuation.

Prior work therefore augments CLIP-score-based filtering with additional image- and text-based signals, such as image similarity to target data, task-relevant labels, or sample uniqueness.
These criteria capture complementary aspects of data utility: alignment reflects pair cleanness, target similarity reflects task relevance, and uniqueness reduces redundancy while improving coverage.
However, they are usually introduced as separate heuristics and combined through manually designed rules, intersections, or weighted mixtures \cite{DataComp, SCLIP, HYPE, CLIPCOV}, which can lead to inconsistent performance across datasets, subset sizes, and evaluation settings.
This motivates a unified and principled framework that explains the role of each criterion in data valuation.

In this work, we revisit data selection for CLIP from an influence-inspired perspective.
Classical influence functions quantify how individual training samples affect a target task, but their direct application to CLIP pretraining is computationally infeasible since it requires per-sample gradients or Hessians \cite{Blackbox, TracIn, Influenceestimation}.
Rather than obtaining classical influence scores, we use the underlying idea of sample-target interaction to derive a scalable data valuation principle for CLIP-style models.

We propose \textsc{InfCLIP}, a computationally efficient scoring method based on temperature-rescaled softmax interactions between training and target samples.
Using only feature representations from a pretrained teacher in a single forward pass, \textsc{InfCLIP} measures how strongly each training sample interacts with the target distribution without retraining or explicit gradient computation.
Importantly, alignment, target relevance, and sample uniqueness naturally emerge within this unified formulation, rather than being manually combined through separate heuristics as in prior methods.
Although \textsc{InfCLIP} is derived as a scalable data valuation score rather than a direct estimator of classical influence, its high-scoring samples align well with classical notions of influence.
In particular, \textsc{InfCLIP} exhibits strong positive correlation with TracIn \cite{TracIn} and Group Shapley value \cite{shapley}, especially among high-scoring samples.

We then apply \textsc{InfCLIP} to practical data selection for CLIP pretraining.
For subset selection, \textsc{InfCLIP} identifies compact subsets that preserve or even surpass the performance of full-data training, consistently outperforming all baseline selection methods across in-distribution and distribution-shifted tasks (Fig.~\ref{fig:pruning}).
We also introduce \textsc{Self-InfCLIP}, a self-influence variant for efficient training-set analysis, including noise detection and memorization-aware assessment, without additional retraining.
Together, these results establish \textsc{InfCLIP} as a unified influence-inspired framework for data selection and dataset diagnosis on large-scale image-text corpora.
Our code is available at \url{\codeurl}.

%% file: sections/2_Related.tex
\paragraph{Data Valuation in CLIP}
Data valuation plays a critical role in CLIP training, where training data are `common crawled' at web scale and inevitably contain noisy, harmful, or redundant samples. Recent progress in multimodal representation learning, in particular CLIP \cite{CLIP}, has highlighted both the potential and the challenges of leveraging large-scale image-text datasets, motivating extensive research on data filtering and valuation.

In contrast to supervised learning--where data valuation has been extensively studied from diverse perspectives, including data difficulty \cite{forgetting, EL2N, CG}, training dynamics \cite{cartography}, influence estimation \cite{TracIn}, and coreset construction \cite{BWS, CRAIG}--data valuation for web-scale multimodal pretraining remains relatively underexplored. 
Most existing approaches rely on the CLIP score, defined as the cosine similarity between image and text embeddings. DataComp \cite{DataComp} and subsequent works combine this with image- and text-based filters or additional signals such as rarity and distributional statistics. Other methods go beyond raw alignment by modeling semantic specificity or redundancy, for example via hyperbolic entailment embeddings (HYPE \cite{HYPE}), regularized alignment (SCLIP \cite{SCLIP}), clustering and deduplication \cite{SemDeDup, EffectivePrune}, multiple filtering signals for downstream performance \cite{ecodatum}, or coreset and coverage objectives \cite{CLIPCOV}. Auxiliary or teacher models have also been used for data selection, including target distribution matching \cite{Filterlikeyoutest}. Related studies further examine memorization in CLIP-style models to identify potentially harmful or overly memorized samples \cite{CLIPMem}.

Despite these advances, most multimodal data valuation methods remain largely heuristic. Alignment-based scores capture semantic consistency but not informativeness, while multi-signal pipelines require careful tuning and lack a unified, principled foundation, motivating the need for a more systematic framework for multimodal data valuation.

\paragraph{Influence Functions}
Assessing the impact of individual data points on model training has long been a central topic in data valuation. Influence functions provide an efficient approximation of this impact by estimating the effect of removing a training sample without requiring full retraining. The influence of a data point $z$ can be quantified by the parameter change induced by its removal, $\hat{\theta}-\hat{\theta}_{-z}$, and the resulting change in test loss, i.e., $\mathrm{Inf}(z) = L_{\mathrm{test}}(\hat{\theta}) - L_{\mathrm{test}}(\hat{\theta}_{-z})$. Influence-based methods have been successfully applied to identifying noisy or mislabeled data, detecting important training examples, and analyzing model behavior \cite{Blackbox, Influenceestimation}.

Despite their theoretical appeal, influence estimation is computationally expensive. For a model with $p$ parameters, gradients and Hessians scale as $\mathcal{O}(p)$ and $\mathcal{O}(p^2)$, making exact computation intractable for modern deep networks. To mitigate this cost, prior work has proposed various approximations, including checkpoint-based gradient similarity \cite{TracIn}, random projections of gradients \cite{TRAK}, layer-wise or block-diagonal Hessian approximations \cite{DataInf}, and other scalable methods \cite{ScalingInf}. 
Even with these approximations, influence-based methods are mostly limited to small or medium-scale datasets, whereas multimodal pretraining corpora such as those used for CLIP are orders of magnitude larger, rendering existing techniques computationally impractical. To address this, we propose \textsc{InfCLIP}, a computationally efficient, influence-inspired scoring method that replaces loss-based influence with a softmax-based formulation.

%% file: sections/3_Method.tex
\subsection{Preliminaries}
\paragraph{CLIP Training} A CLIP model consists of an image encoder $F(\cdot)$ and a text encoder $G(\cdot)$.
For an image-text pair $(x_i, y_i)$, let the corresponding embeddings be $f_i = F(x_i)$ and $g_i = G(y_i)$. The model is trained with the symmetric InfoNCE loss \cite{infonce}:
\begin{equation*}
-\dfrac{1}{N}\sum\limits_{i=1}^N \Big[
\log \frac{\exp(f_i^\top g_i / \tau)}{\sum_k \exp(f_i^\top g_k / \tau)} +
\log \frac{\exp(f_i^\top g_i / \tau)}{\sum_k \exp(f_k^\top g_i / \tau)}
\Big],
\end{equation*}
where $\tau>0$ is a temperature parameter. This objective encourages high similarity between matched image-text pairs while discouraging mismatched ones.
For notational convenience, we define the \emph{softmax score} of data by sum of image-to-text and text-to-image softmax probabilities, denoted by: 
\begin{equation}\label{eq:softmax}
s_i = p_{ii} + q_{ii} \;\;\text{ where }
p_{ij} = \frac{\exp(f_i^\top g_j / \tau)}{\sum_k \exp(f_i^\top g_k / \tau)}, \,\, 
q_{ij} = \frac{\exp(f_i^\top g_j / \tau)}{\sum_k \exp(f_k^\top g_j / \tau)}.
\end{equation}

\paragraph{Influence Functions}
Influence functions \cite{Blackbox} quantify how an individual training sample affects model parameters and predictions on target data. A common approximation of the influence of sample $i$ on a target sample $t$ can be derived via a first-order Taylor expansion. Let $W$ denote the model parameters trained on the full dataset and $W_{-i}$ those obtained when the $i$-th sample is removed. Then,
\begin{equation*}
\begin{aligned}
f_{W_{-i}}(t) - f_W(t)
&\simeq (W_{-i} - W) \cdot \nabla_W f(t) \simeq \eta\nabla_W \ell(i) \cdot \nabla_W f(t)
\end{aligned}
\end{equation*}
In this work, rather than estimating the parameter change caused by removing a sample, we adopt an influence-inspired perspective based on sample-target interactions in representation space.
Specifically, we use gradients of the softmax score $s_i$ in \eqref{eq:softmax}, which directly reflects the image-text matching behavior optimized by CLIP.

\subsection{InfCLIP Score: Formulation and Approximation}\label{sec:definition}
We quantify the influence between a training sample $i$ and a target sample $t$ as the inner product of their softmax score gradients with respect to the model parameters, which quantifies how a change in the softmax score of sample $i$ affects that of sample $t$. Formally, the influence score is defined as
\begin{equation*}
\nabla_W s_i \cdot \nabla_W s_t
= \nabla_F s_i \cdot \nabla_F s_t
+ \nabla_G s_i \cdot \nabla_G s_t,
\end{equation*}
where $W=(F,G)$ denotes the model parameters, and $\nabla_F$ and $\nabla_G$ denote gradients with respect to the image and text encoders.

\paragraph{Gradient Structure under Softmax}
Recall that for a softmax variable
$\hat{z}_i = \exp(z_i) / \sum_k \exp(z_k)$, the derivatives satisfy
$\partial \hat{z}_i / \partial z_i = \hat{z}_i (1 - \hat{z}_i)$ and
$\partial \hat{z}_i / \partial z_j = -\hat{z}_i \hat{z}_j$ for $j \neq i$.
Using these identities, the gradients of the image-to-text and text-to-image
probabilities, $p_{ii}$ and $q_{ii}$ in \eqref{eq:softmax}, with respect to the image encoder parameters
can be written as

\begin{equation*}
\nabla_F p_{ii}
= \frac{p_{ii}}{\tau} J_i^\top \bar{g}_i,
\qquad
\nabla_F q_{ii}
= \frac{q_{ii}}{\tau} \bar{J}_i^\top g_i,
\end{equation*}
where $\bar{g}_i := g_i - \sum_k p_{ik} g_k$ denotes the centered text embedding,
$J_i = \nabla_F f_i \in \mathbb{R}^{d \times |F|}$ is the Jacobian of the image
encoder output, and
$\bar{J}_i := \nabla_F f_i - \sum_k q_{ki} \nabla_F f_k$ is the centered
Jacobian.
Analogous expressions hold for gradients with respect to the text encoder.

Combining the two softmax terms gives
\begin{equation*}
\nabla_F s_i \cdot \nabla_F s_t
=
\left(
\frac{p_{ii}}{\tau}J_i^\top\bar{g}_i
+
\frac{q_{ii}}{\tau}\bar{J}_i^\top g_i
\right)
\cdot
\left(
\frac{p_{tt}}{\tau}J_t^\top\bar{g}_t
+
\frac{q_{tt}}{\tau}\bar{J}_t^\top g_t
\right),
\end{equation*}
which expands into four terms, e.g., $p_{ii}p_{tt}\bar{g}_i^\top J_iJ_t^\top\bar{g}_t/\tau^2$.
The text-encoder contribution has the same form with $\bar{f}_i := f_i-\sum_k q_{ki}f_k$, $K_i:=\nabla_G g_i$, and $\bar{K}_i:=\nabla_G g_i-\sum_k p_{ik}\nabla_G g_k$.

\paragraph{Approximation via an NTK Perspective}
Explicitly computing Jacobians and their inner products is computationally prohibitive for large-scale models and datasets.
We therefore approximate these Jacobian interactions from a Neural Tangent Kernel (NTK) perspective \cite{NTK}.
For $J_i = \nabla_F f_i \in \mathbb{R}^{d \times |F|}$, the matrix
$J_i J_t^\top \in \mathbb{R}^{d \times d}$ corresponds to the output-space NTK
between images $x_i$ and $x_t$.
Although this NTK is generally matrix-valued for vector-valued networks, theoretical analyses of wide neural networks show that, under standard assumptions such as symmetric initialization and approximate isotropy across output channels, it concentrates around a scalar kernel times the identity.
To obtain a tractable feature-based score without evaluating gradients, we further approximate the scalar kernel using the learned representation similarity, following common practice in random feature/kernel methods and scalable influence estimation \cite{RandomFeatureNTK, learningwithkernels, TracIn, BWS}:
\begin{equation*}
J_i J_t^\top \approx k_F(x_i,x_t) I_d,
\qquad
k_F(x_i,x_t) \approx f_i^\top f_t .
\end{equation*}

This feature-product structure can also be derived from a last-layer perspective by restricting gradients to the final linear projection layers of CLIP-style models.
We provide this alternative derivation in Appendix~\ref{sec:lastlayer}.

Substituting the NTK approximation into all gradient inner products yields

{\small
\begin{equation}
\label{eq:infclip_score}
I(i,t)
=
\frac{2}{\tau^2}\Big[
p_{ii}p_{tt}(f_i^\top f_t)(\bar{g}_i^\top \bar{g}_t)
+
p_{ii}q_{tt}(\bar{g}_i^\top g_t)(f_i^\top \bar{f}_t)
+
q_{ii}p_{tt}(\bar{f}_i^\top f_t)(g_i^\top \bar{g}_t)
+
q_{ii}q_{tt}(\bar{f}_i^\top \bar{f}_t)(g_i^\top g_t)
\Big]
\end{equation}
}

Each term is the product of softmax weights and feature similarities between adjusted image/text representations.
Thus, \textsc{InfCLIP} makes explicit how alignment, target relevance, and uniqueness interact in a single influence-inspired score.

\paragraph{Temperature Rescaling}
In practice, the softmax distributions induced by pretrained CLIP models are often overly concentrated, leading to vanishing gradients and weak influence signals.
Specifically, when $p_{ii} \approx 1$ and $p_{ik} \approx 0$ for $k \neq i$, the softmax expectation $\sum_k p_{ik} g_k$ collapses to $g_i$, yielding a nearly zero centered embedding vector $\bar{g}_i = g_i - \sum_k p_{ik} g_k \approx 0$ and hence a negligible gradient $\nabla_W s_i$.
This issue arises because the scores are computed using features from a fully converged model, where gradients are already saturated and no longer informative about sample interactions.
Accordingly, prior influence-function methods often rely on intermediate checkpoints or training dynamics, where gradients remain meaningful and per-sample effects are more distinguishable \cite{TracIn, CapturingTempInf, WhatInfdo}.
Rather than modifying the model parameters or retraining at earlier epochs, we address this issue by applying \emph{temperature rescaling} only to the softmax normalization used in InfCLIP score computation (i.e., $p_{ij}$, $q_{ij}$, $\bar{g}_i$, and $\bar{f}_i$).
This preserves the semantic information encoded in the pretrained features while preventing the softmax from becoming nearly one-hot, thereby maintaining informative gradients without altering the underlying CLIP representations.
Importantly, this rescaling affects only the scoring procedure and does not modify the pretrained CLIP model itself.
In our experiments, we extract features using a pretrained CLIP model with temperature $\tau_{\mathrm{CLIP}}=0.01$ and compute \textsc{InfCLIP} scores with a rescaled temperature $\tau_{\mathrm{score}}=0.1$.
\textsc{InfCLIP} is robust to the choice of temperature and batch size; ablation studies on these hyperparameters are provided in Appendix~\ref{sec:Ablation}.

\subsection{Comparison with Classical Influence Scores}

\begin{wrapfigure}{r}{0.50\linewidth}
\vspace{-4mm}
    \centering
    \includegraphics[width=0.48\linewidth]{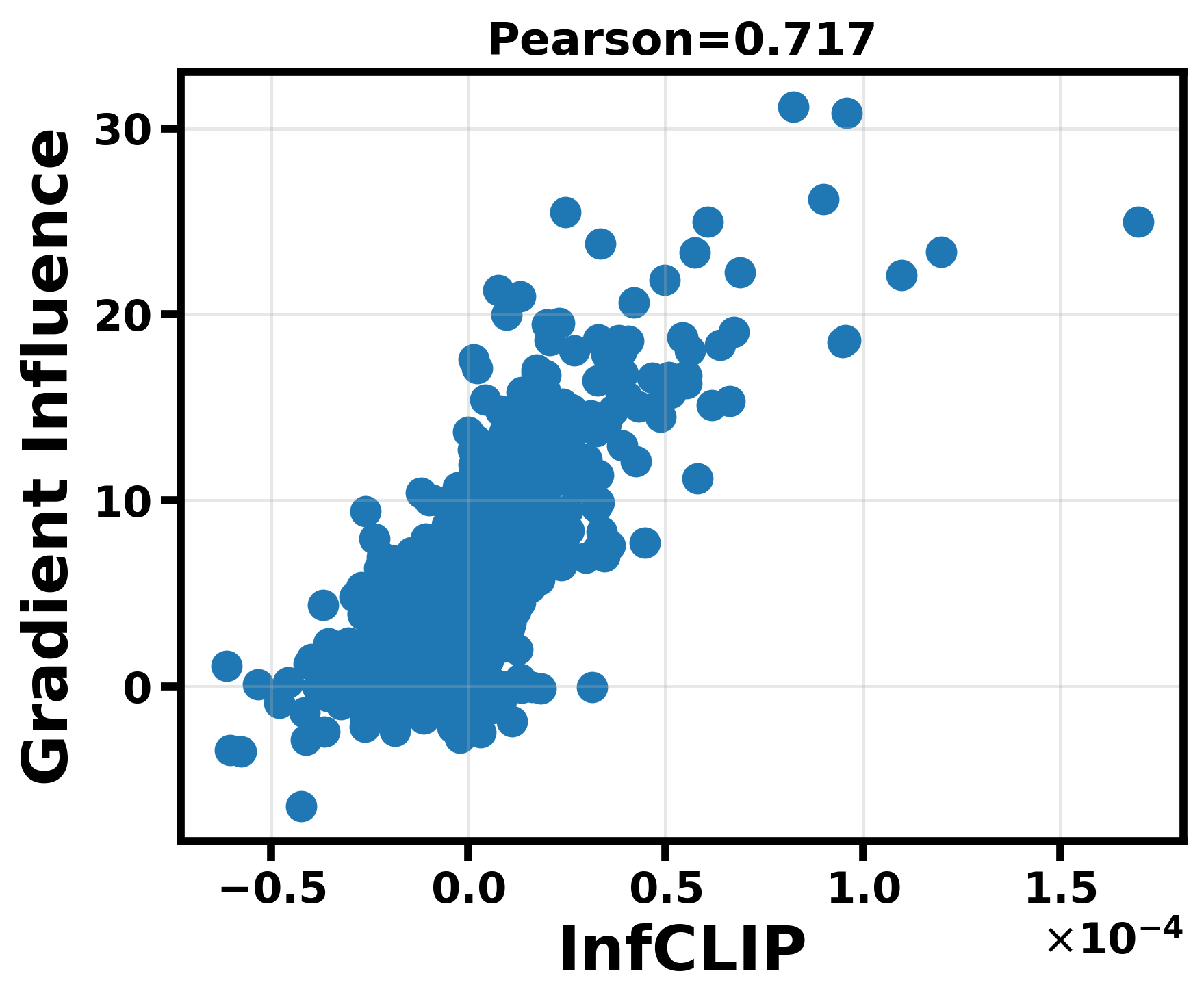}
    \hfill
    \includegraphics[width=0.48\linewidth]{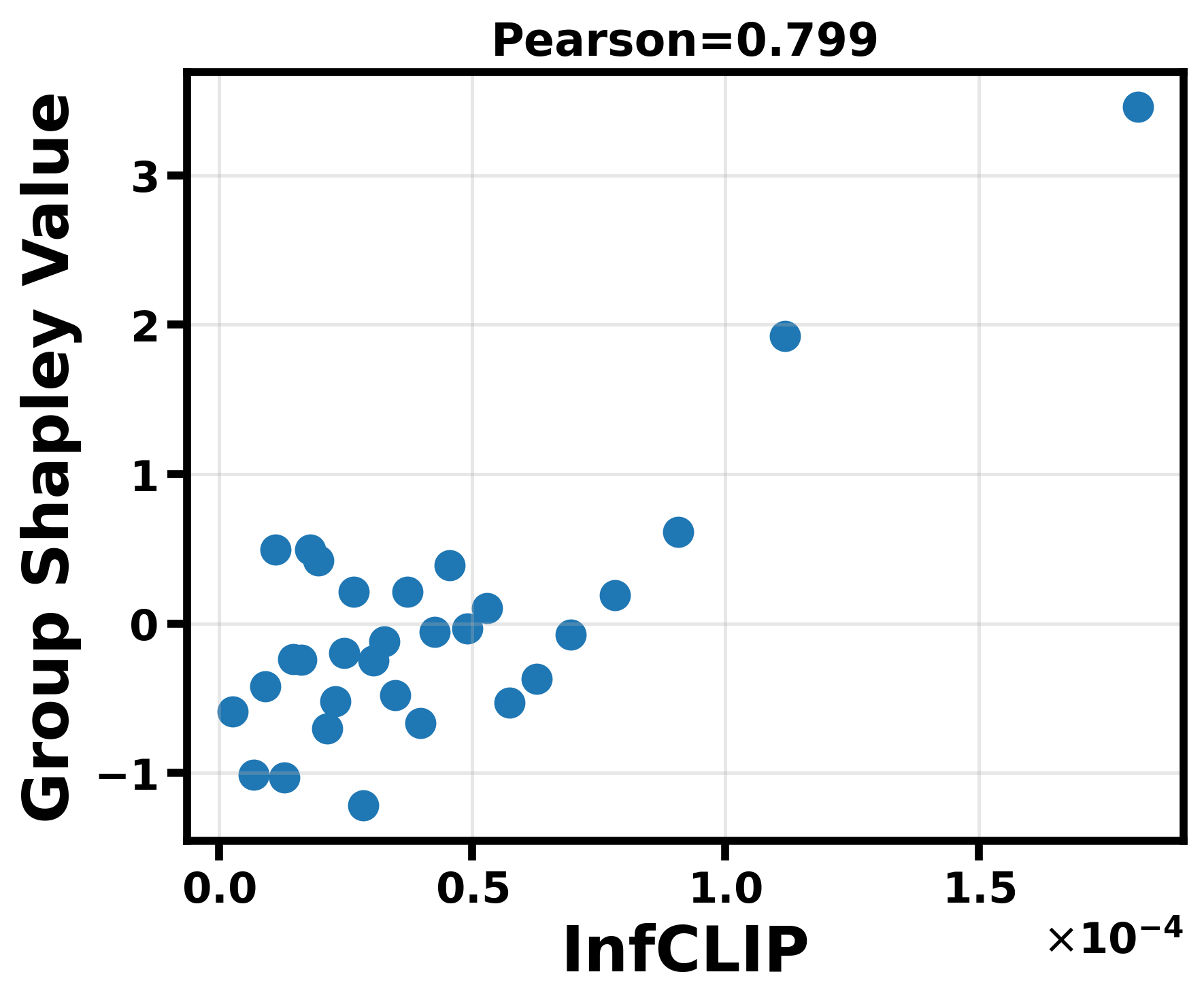}
    \caption{\small Scatter plots of \textsc{InfCLIP} with TracIn (left) and Group Shapley (right).}
    \label{fig:influence_consistency}

    \vspace{-1mm}

    \captionof{table}{\small Reverse pruning results.}
    \label{tab:reversepruning}
    \resizebox{\linewidth}{!}{
    \begin{tabular}{c|cccccc}
        \toprule
        Prune (\%) & 0 & 1 & 3 & 5 & 10 & 20 \\
        \midrule
        InfCLIP & 16.37 & \textbf{14.93} & \textbf{13.26} & \textbf{12.03} & \textbf{9.17} & \textbf{6.17} \\
        CLIP & 16.37 & 16.05 & 15.94 & 15.41 & 14.85 & 12.93 \\
        Random & 16.37 & 15.87 & 15.69 & 15.55 & 15.04 & 14.58 \\
        \bottomrule
    \end{tabular}}
\vspace{-4mm}
\end{wrapfigure}

Although \textsc{InfCLIP} is not designed to explicitly approximate classical influence scores, we verify that it is highly correlated with classical influence measures.
We compare \textsc{InfCLIP} with TracIn \cite{TracIn}, which measures sample-target influence by the gradient inner product $\mathrm{TracIn}(i,t)=\nabla_{\theta}\ell(i)\cdot\nabla_{\theta}\ell(t)$, and the Shapley value \cite{shapley}, which measures the expected marginal contribution of a sample to a performance utility.
Formally, the Shapley value is defined as $\phi_i=\mathbb{E}_{S\subset D\setminus\{i\}}[U(S\cup\{i\})-U(S)]$, where $U(S)$ denotes the downstream performance of a model trained on subset $S$.
Since exact sample-wise Shapley estimation is infeasible at CLIP scale, we estimate group-level Shapley values by sorting CC3M samples according to \textsc{InfCLIP}, partitioning them into 30 score-ranked groups, and estimating each group's marginal contribution via Monte Carlo.

Classical influence measures are computationally expensive at scale.
Shapley estimation requires training many models from scratch on different subsets, while TracIn requires backpropagation and storage of per-sample gradients.
In contrast, \textsc{InfCLIP} only requires a single forward pass to obtain image and text features, followed by score computation.
As shown in Table~\ref{tab:time}, this takes less than one minute on CC3M, and the memory cost is limited to storing feature representations rather than parameter gradients.
Experimental details are provided in Appendix~\ref{sec:exp_details}.

Figure~\ref{fig:influence_consistency} shows strong positive correlations, especially in the high-scoring regime.
This indicates that samples assigned high scores by \textsc{InfCLIP} are also regarded as influential by classical influence measures.
Therefore, \textsc{InfCLIP} provides a practical influence-inspired surrogate that avoids repeated retraining and explicit gradient computation.

We further examine whether these high-\textsc{InfCLIP} samples are essential for CLIP's zero-shot performance.
To this end, we conduct a reverse pruning experiment on CC3M: we progressively remove the top-scoring samples according to each method, retrain CLIP models from scratch on the reduced datasets, and evaluate zero-shot performance on ImageNet.
Table~\ref{tab:reversepruning} compares pruning based on \textsc{InfCLIP}, CLIP score, and random removal.
Removing high-\textsc{InfCLIP} samples leads to a substantially faster performance degradation than the two baselines.
In particular, discarding only the top 10\% of samples by \textsc{InfCLIP} reduces zero-shot accuracy from 16.37\% to 9.17\%.
This shows that the high-scoring samples, where \textsc{InfCLIP} aligns well with classical influence measures, are also critical for CLIP pretraining.
We further present representative examples ranked by \textsc{InfCLIP} in Appendix~\ref{sec:Real_examples}, which show that high-scoring samples typically exhibit strong semantic alignment, relevance to ImageNet concepts, and distinctive visual or textual content.

\subsection{Influential Data: A Unified Measure}

The \textsc{InfCLIP} score $I(i,t)$ in \eqref{eq:infclip_score} decomposes into terms consisting of
(i) softmax-based alignment weights $(p_{ii},p_{tt},q_{ii},q_{tt})$ and
(ii) interactions between original and centered image/text representations, such as $f_i,\bar f_i,g_i,\bar g_i$.
This decomposition provides a unified view of several criteria that have been used separately in prior methods: alignment, target relevance, and sample uniqueness.

\paragraph{Alignment}
Alignment-based filtering is the most common criterion for CLIP data selection.
CLIP score uses the image-text similarity $f_i^\top g_i$ to measure whether an image and its caption are semantically matched, while SCLIP \cite{SCLIP} refines this idea through a softmax-based alignment score $p_{ii}+q_{ii}$.
In \textsc{InfCLIP}, alignment appears through the multiplicative softmax weights $p_{ii}$ and $q_{ii}$.
Rather than treating alignment as a separate selection criterion, \textsc{InfCLIP} naturally incorporates it as a multiplicative weight on the feature-interaction terms, allowing alignment to be considered jointly with other criteria.

\paragraph{Target relevance via intra-modal similarity}
A second line of methods selects samples that are relevant to the target distribution.
On the image side, similarity-based methods such as NormSim \cite{SCLIP} use visual feature similarity to favor samples close to target images.
On the text side, DataComp-style label or keyword filtering selects captions containing target-relevant concepts \cite{DataComp}.
\textsc{InfCLIP} captures both principles through intra-modal feature interactions: terms such as $f_i^\top f_t$ and $\bar f_i^\top \bar f_t$ measure image-side relevance, while $g_i^\top g_t$ and $\bar g_i^\top \bar g_t$ provide the analogous text-side relevance.
Therefore, target relevance is incorporated directly into the \textsc{InfCLIP}, without requiring manually specified labels, keywords, or separate similarity filters.

\paragraph{Uniqueness via centered representations}
Another important criterion is sample uniqueness, which aims to avoid selecting redundant examples from dense regions of the data distribution.
SemDeDup removes near-duplicate or semantically redundant samples using embedding-space similarity, while HYPE emphasizes specific and informative samples by measuring image and text specificity \cite{SemDeDup, HYPE}.
In \textsc{InfCLIP}, uniqueness naturally arises from centered representations such as
$\bar g_i=g_i-\sum_k p_{ik}g_k$, where $\sum_k p_{ik}g_k$ is the expected text feature conditioned on image $x_i$.
The centered feature $\bar g_i$ measures how much the given caption deviates from the model's expected description.
Distinctive samples tend to have large $\|\bar g_i\|$, whereas redundant samples in dense regions have smaller centered-feature norms.

\begin{figure}[t]
\begin{minipage}[t]{0.45\textwidth}
\vspace{-1in}
    \centering
    \captionof{table}{\small Rank correlations between individual components of the \textsc{InfCLIP} score (rows) and existing data valuation measures (columns).}
    \resizebox{\linewidth}{!}{
    \begin{tabular}{c|c|c|c|c}
    \toprule
        & $\mathrm{CLIP}$ 
        & $\mathrm{NormSim}_f$ 
        & $\mathrm{Uniq}_g$ 
        & $\mathrm{\textsc{InfCLIP}}$ \\ 
    \midrule
        $p_{ii}$ 
        & \textbf{0.921} & -0.090 & 0.460  & 0.766 \\ 
    \midrule
        $\cos(\bar{f}_i, \bar{f}_t)$ 
        & -0.043 & \textbf{0.554} & -0.017 & 0.438 \\ 
    \midrule
        $\|\bar{g}_i\|$ 
        & 0.391 & 0.003 & \textbf{0.906}  & 0.384 \\ 
    \midrule
        $\mathrm{\textsc{InfCLIP}}$ 
        & 0.683 & 0.253 & 0.431 & 1.000 \\ 
    \bottomrule
    \end{tabular}}
    \label{tab:Correlations}
\end{minipage}
\hfill
\begin{minipage}[t]{0.50\textwidth}
    \centering
    \includegraphics[width=0.99\linewidth]{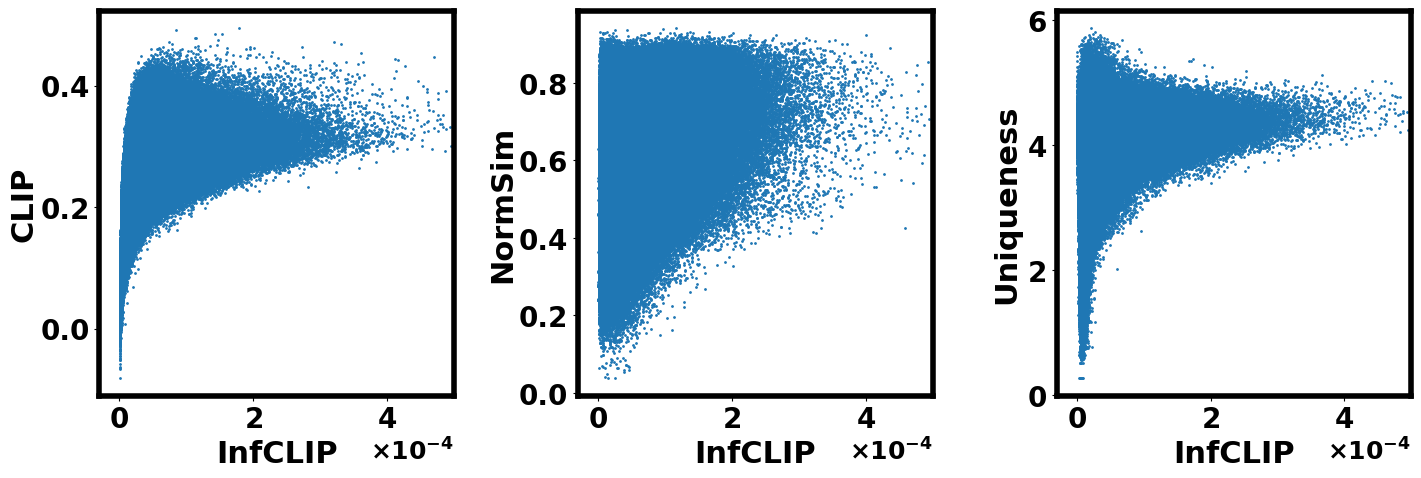}
    \caption{\small Scatter plots of the \textsc{InfCLIP} score against CLIP, NormSim, and Uniqueness.}
    \label{fig:Scatter}
\end{minipage}
\end{figure}

\paragraph{Empirical verification of unified criteria}
To empirically verify that \textsc{InfCLIP} subsumes these existing criteria all together, we compute representative scores for each aspect: CLIP score for alignment, NormSim for image-side target relevance, and the kNN-based uniqueness score $\mathrm{Uniq}_g$ for sample uniqueness. 
For the uniqueness score, we obtain $N_k(i)$, the set of $k$ nearest neighbors of $g_i$ in the text feature space, and define
$\mathrm{Uniq}_g(i):=\sum_{j\in N_k(i)}\|g_i-g_j\|_2$.
This score is large for samples in low-density regions and small for redundant samples in dense regions.
Table~\ref{tab:Correlations} reports Spearman rank correlations between these scores and the corresponding components of \textsc{InfCLIP}, evaluated on CC3M trainset with ImageNet targets \cite{CC3M, ImageNet}.
Each \textsc{InfCLIP} component strongly correlates with its matched criterion: $p_{ii}$ with CLIP alignment, $\cos(\bar f_i,\bar f_t)$ with NormSim, and $\|\bar g_i\|$ with $\mathrm{Uniq}_g$.
Meanwhile, correlations among unrelated criteria remain much weaker, showing that these components capture distinct and complementary signals.
Thus, alignment, target relevance, and uniqueness are independently useful criteria, and \textsc{InfCLIP} integrates them into a single influence-inspired score.
Figure~\ref{fig:Scatter} further illustrates that high \textsc{InfCLIP} scores arise only when all three aspects are simultaneously large, whereas a high value in any single component alone is insufficient.
This shows that our InfCLIP score captures multiple aspects of data utility, including alignment, target relevance, and uniqueness, within a unified and effective framework, whereas existing scores typically reflect only a subset of these criteria.

%% file: sections/4_Experiment.tex
\subsection{Subset Selection for CLIP Training}\label{sec:SubsetSelection}

In this section, we evaluate \textsc{InfCLIP} as a data selection criterion for CLIP pretraining, with the goal of assessing its ability to identify high-value training samples that are most beneficial for a given target task. We use the Conceptual Captions 3M/12M (CC3M/12M) \cite{CC3M, CC12M} datasets for pretraining and ImageNet \cite{ImageNet} as the target dataset.
We further evaluate generalization to the DataComp-small
training corpus and investigate the effect of extending the
target set to additional downstream distributions.

\paragraph{Computing InfCLIP Score}
We compute the \textsc{InfCLIP} score for each training sample $i$ with respect to a target dataset $T$ as
\begin{equation*}
    I(i,T) = \max_{t \in T} I(i,t),
\end{equation*}
where $I(i,t)$ denotes the \textsc{InfCLIP} score between training sample $i$ and target sample $t$.
To reduce memory and computational cost, we do not use the full ImageNet dataset as $T$. Instead, we construct a compact \emph{prototype set}. Let $C$ denote the set of ImageNet classes. For image features, we compute class-wise mean embeddings $f_c = \frac{1}{|c|}\sum_{i \in c} f_i$ for each $c \in C$, yielding $|C|$ image prototypes (1,000 for ImageNet). For text features, we apply standard CLIP prompt templates to each class name and compute the corresponding text embedding, resulting in one text prototype per class. The paired image and text prototypes for each class form the target set $T$, so that $|T| = |C|$.
This prototype construction substantially reduces the number of target samples considered during scoring, making data valuation more computationally efficient. To examine the sensitivity to prototype construction, we also consider alternative strategies, including randomly selecting a sample or choosing the most aligned sample from each class.
These alternatives yield similar downstream performance, suggesting that \textsc{InfCLIP} is relatively robust to the choice of target prototypes.
Finally, we ablate key scoring hyperparameters---the softmax temperature and the batch size used to compute $I(i,t)$---with details reported in Appendix~\ref{sec:Ablation}.

\paragraph{Experimental Setup}
We compute data selection scores using normalized features from an OpenCLIP teacher model pretrained on DataComp-XL \cite{DataComp}, except for HYPE, for which we follow the official implementation and use its released reference teacher model \cite{HYPE}.
This may slightly disadvantage HYPE, as its released reference teacher is weaker than DataComp-XL, but we keep the official setting for faithful reproduction.

For subset-selection experiments, we train CLIP models from scratch with a modified ResNet-50 visual encoder and a Transformer-based text encoder.
We use a fixed training budget across all selection methods and subset ratios, corresponding to 10 full-data epochs on CC3M and 5 on CC12M.
For a subset ratio $r$, we train on the selected subset for approximately $1/r$ times as many epochs as the full dataset, ensuring the same total number of optimization steps and sample presentations across all settings.
We use AdamW with batch size 1,024, learning rate $5\times10^{-4}$, and weight decay $0.1$.
For each method, we select the top-scoring samples at each subset ratio and evaluate zero-shot classification accuracy on ImageNet, six ImageNet distribution-shift benchmarks, and seven VTAB-Natural tasks, as well as image--text retrieval performance on COCO \cite{COCO}.
We report the mean and standard deviation over three independent training runs, with detailed task-wise results provided in the Appendix \ref{app:full_result}

\newpage
\paragraph{Baselines}
We compare \textsc{InfCLIP} with representative data selection methods.

\noindent\textbf{(1) CLIP score:} ranks samples by cross-modal alignment $f_i^\top g_i$, a widely used proxy for cleanness.\\
\noindent\textbf{(2) Image filtering:} ranks samples by target-aware image similarity $\max_{t\in T} f_i^\top f_t$, where $T$ denotes the ImageNet target set.\\
\noindent\textbf{(3) SCLIP+NormSim \cite{SCLIP}:} first selects the top-$\alpha\%$ samples by the SCLIP score $p_{ii}+q_{ii}$ in \eqref{eq:softmax}, and then reranks them by NormSim, $\|f_i^\top f(X_{\mathrm{target}})\|_p$; we report the better result among $\alpha\in\{30,50\}$.\\
\noindent\textbf{(4) HYPE+NormSim \cite{HYPE}:} first selects samples by the HYPE score $\mathrm{HYPE}_{\mathrm{score}}=\epsilon_i+\epsilon_t-d_{\mathcal{L}}+\cos(\theta)+c_{\mathrm{IN}}$, where $\epsilon_i,\epsilon_t$ are image/text specificity scores, $d_{\mathcal{L}}$ is a hyperbolic distance term, $\cos(\theta)$ is image-text alignment, and $c_{\mathrm{IN}}$ is an ImageNet-class indicator, and then reranks them by NormSim.\\
\noindent\textbf{(5) CLIPCOV \cite{CLIPCOV}:} selects $S^\ast \in \arg\max_{S\subseteq V,\ |S|\le n_s} F_{\mathrm{ClipCov}}(S)$, where $F_{\mathrm{ClipCov}}(S):=F_{\mathrm{cov}}(S)+F_{\mathrm{label}}(S)-F_{\mathrm{class}}^{\mathrm{reg}}(S)+F_{\mathrm{inter}}(S)$, manually combining covariance, label/class coverage, regularization, and interaction terms.

\begin{figure}[t]
    \centering
    \includegraphics[width=0.99\linewidth]{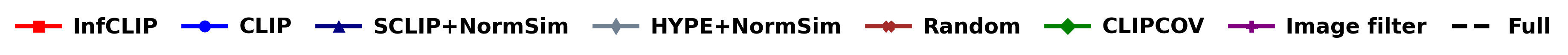}

    \vspace{1mm}

    \begin{subfigure}[b]{0.99\linewidth}
        \centering
        \includegraphics[width=0.32\linewidth]{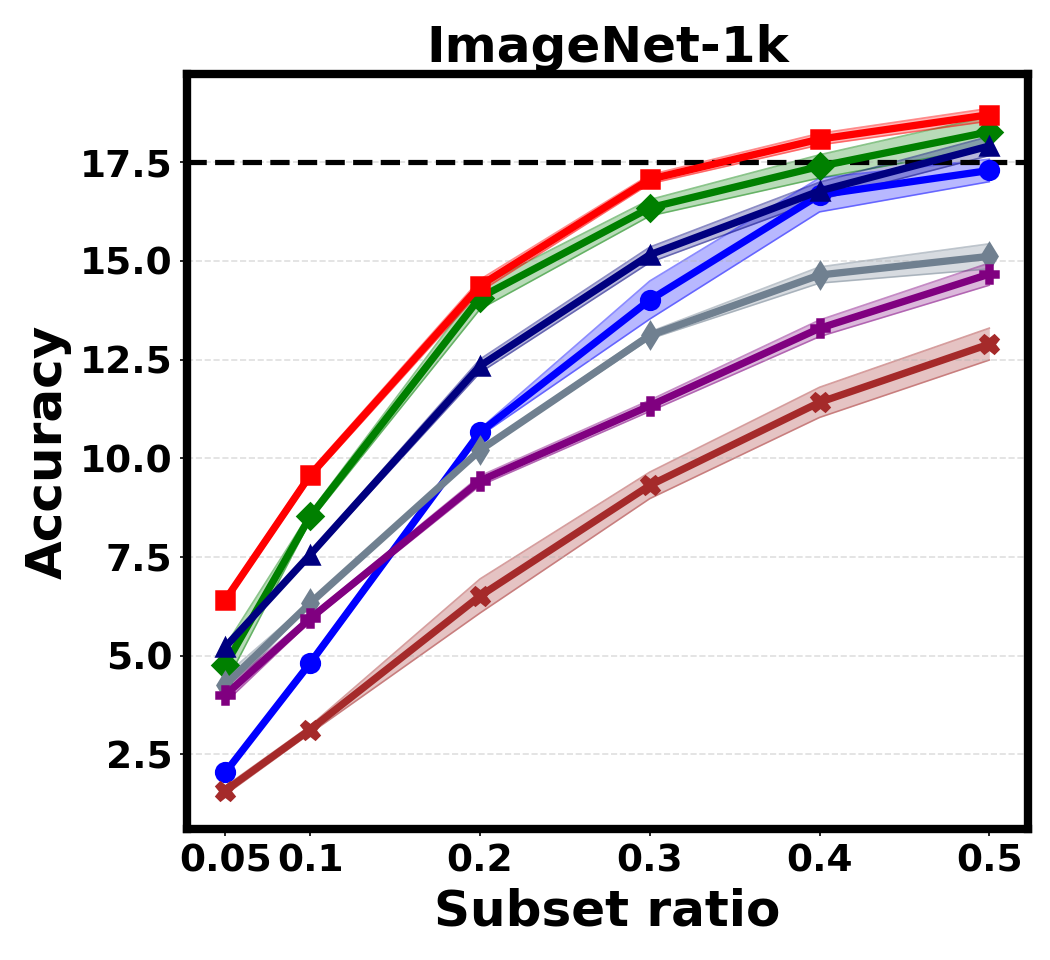}
        \includegraphics[width=0.32\linewidth]{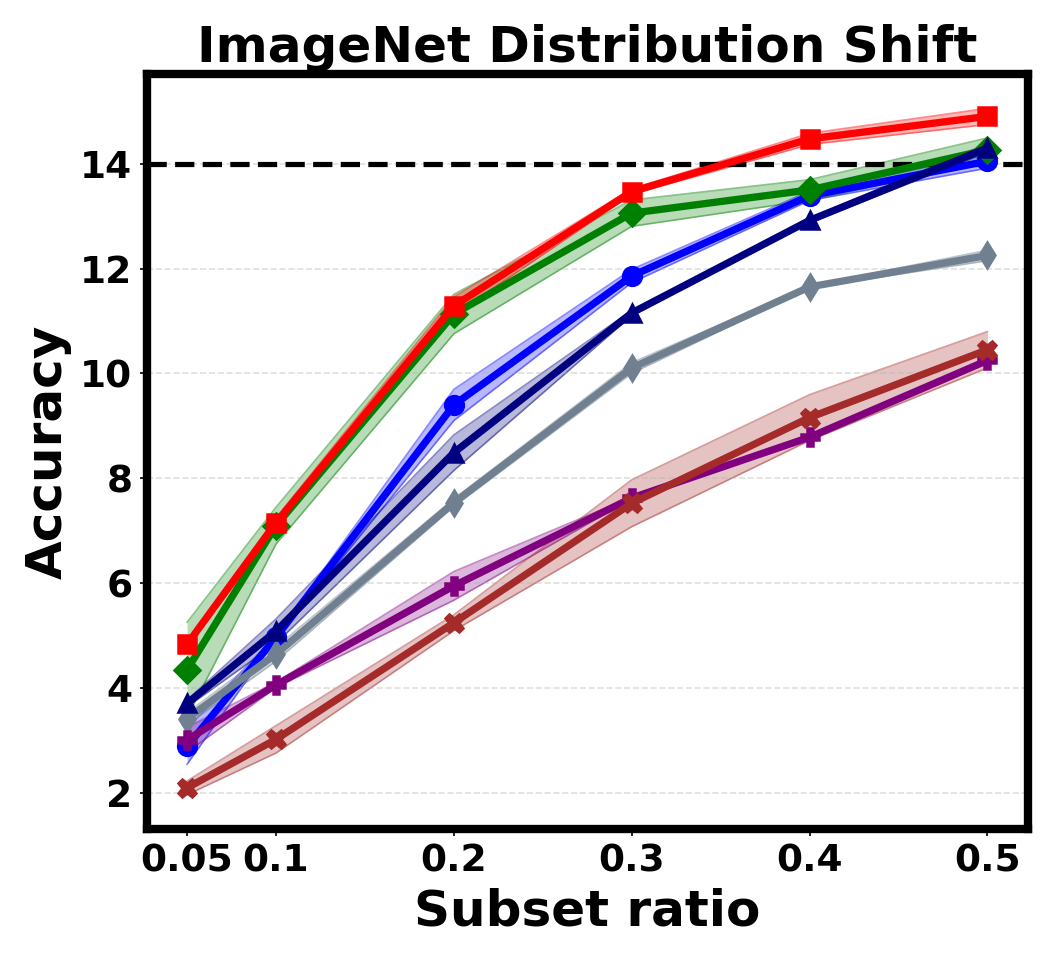}
        \includegraphics[width=0.32\linewidth]{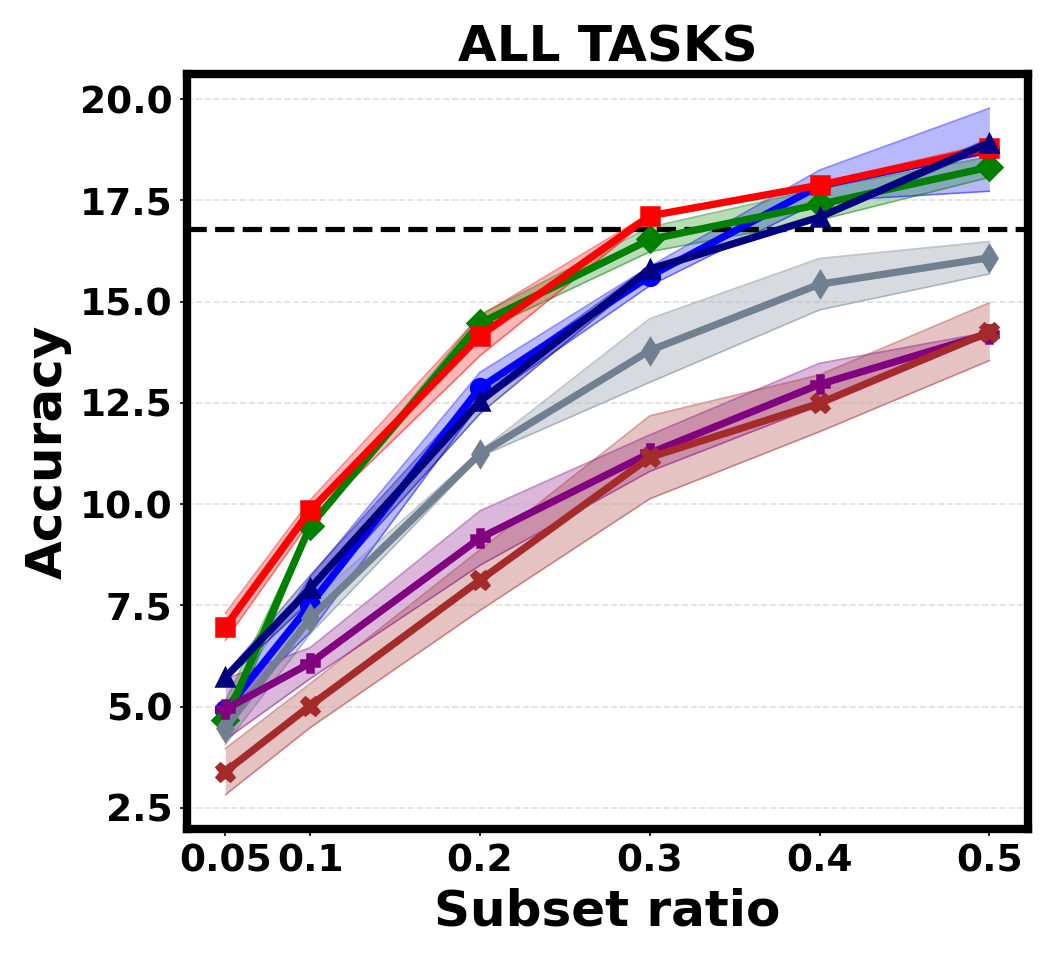}
        \caption{Subset selection on \textbf{CC3M} dataset. The x-axis and y-axis denote subset ratio and accuracy, respectively.}
        \label{fig:pruning_cc3m}
    \end{subfigure}

    \vspace{1mm}

    \begin{subfigure}[b]{0.99\linewidth}
        \centering
        \includegraphics[width=0.32\linewidth]{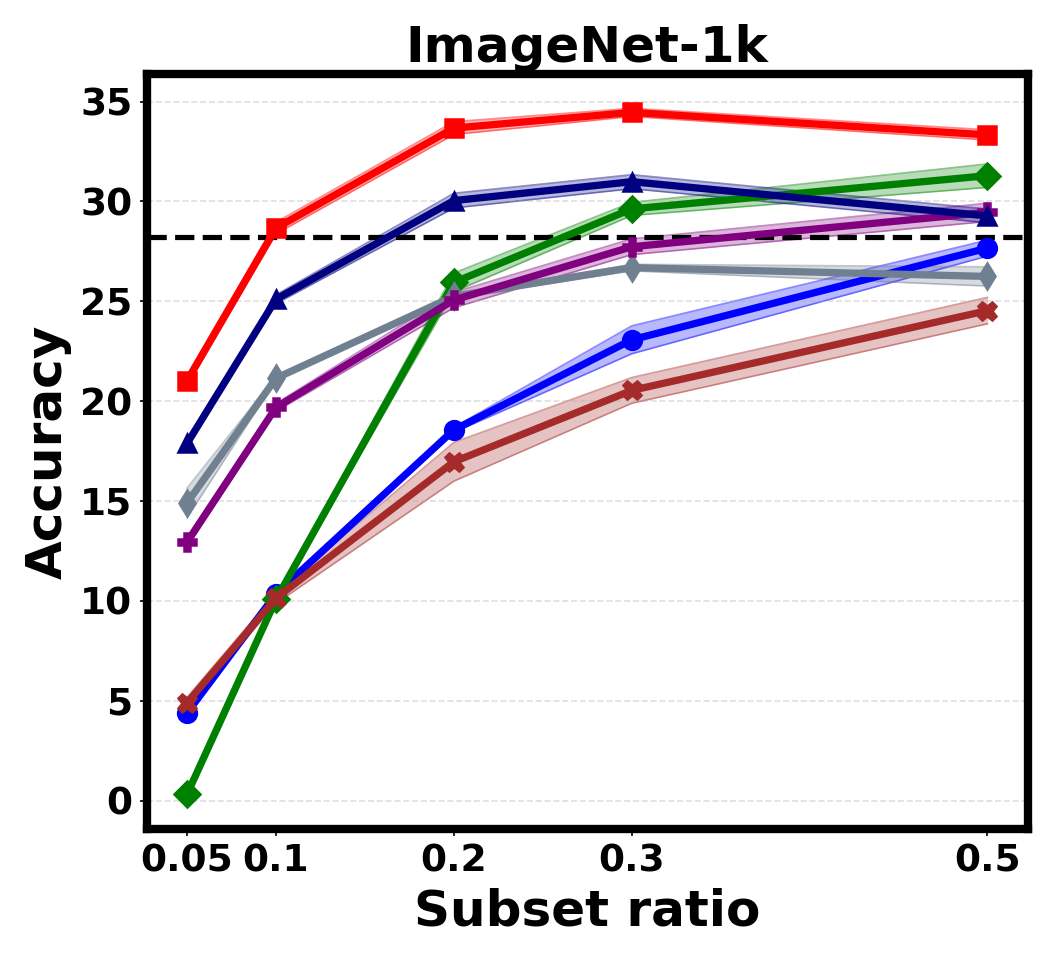}
        \includegraphics[width=0.32\linewidth]{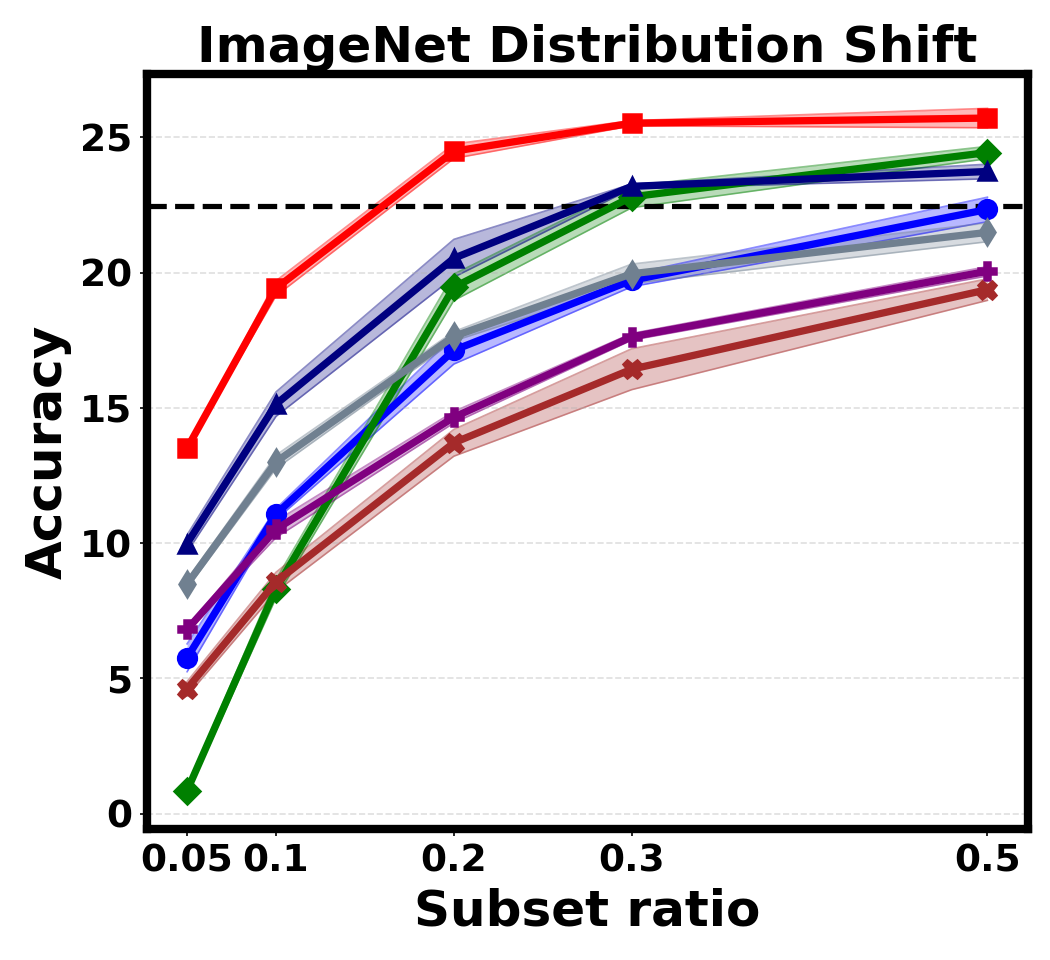}
        \includegraphics[width=0.32\linewidth]{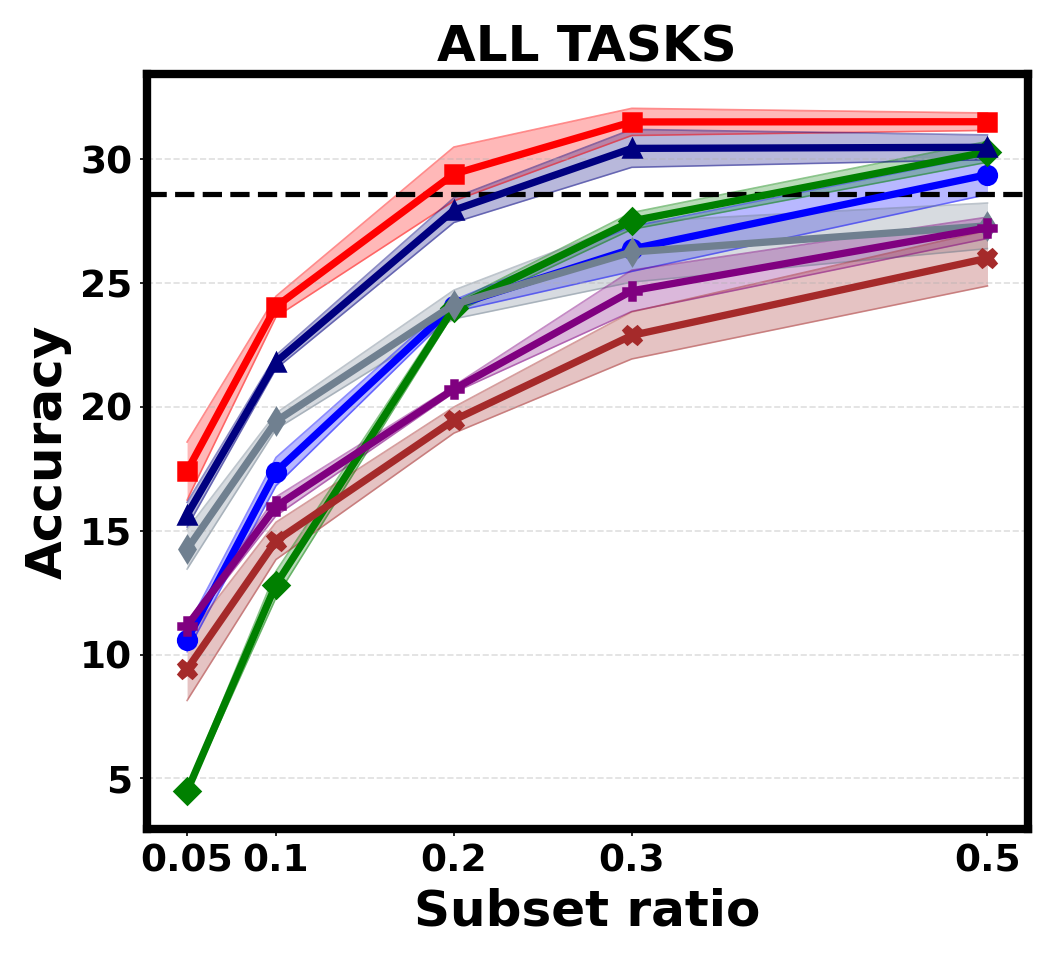}
        \caption{Subset selection on \textbf{CC12M} dataset. The x-axis and y-axis denote subset ratio and accuracy, respectively.}
        \label{fig:pruning_cc12m}
    \end{subfigure}
    
    \caption{Zero-shot performance of CLIP models trained on selected subsets with varying subset ratios.
    We report results on CC3M (a) and CC12M (b), each evaluated on ImageNet-1k (left), distribution-shifted ImageNet benchmarks (middle), and all tasks including VTAB-Natural tasks and COCO retrieval dataset (right).
    \textsc{InfCLIP} consistently outperforms the baselines across datasets and evaluation suites.
    }
    \label{fig:pruning}
\end{figure}

\paragraph{Results}
Figure~\ref{fig:pruning} reports the zero-shot accuracy of models trained on subsets of CC3M/12M (subset ratio on the x-axis), selected by different scoring methods, evaluated on three benchmarks: ImageNet-1k, the average over six ImageNet distribution-shift datasets, and the average over all tasks (ImageNet-1k, six distribution shifts, seven VTAB-Natural tasks, and COCO retrieval dataset). \textsc{InfCLIP} consistently outperforms all baselines across evaluation benchmarks and a wide range of subset ratios. 
Notably, with only 30\%--40\% of the CC3M data or 10\%--20\% of CC12M data, \textsc{InfCLIP} already surpasses the performance of the full-dataset baseline on all three benchmarks, demonstrating that carefully selected subsets can outperform training on the entire dataset. 
Moreover, \textsc{InfCLIP} achieves strong performance on distribution-shifted ImageNet and on the aggregate ``All Tasks'' benchmark, indicating robust generalization beyond the specific target set (e.g., ImageNet-1k) used to compute the scores. Unlike baseline methods that combine alignment, target relevance, and uniqueness through multi-stage filtering or complex objective optimization, \textsc{InfCLIP} relies on a single unified score that jointly captures these factors through an influence-grounded formulation.

We further evaluate \textsc{InfCLIP} on DataComp-small, a noisier and more diverse web-scale dataset.
\textsc{InfCLIP} substantially outperforms CLIP score and SCLIP+NormSim across all evaluated subset ratios and benchmarks, with particularly large gains at smaller selection budgets, demonstrating its effectiveness beyond the CC3M/CC12M datasets.
We also investigate target-set expansion using COCO and VOC2007\cite{voc2007}, showing that incorporating additional target distributions improves performance on the corresponding downstream tasks while largely preserving ImageNet accuracy.
Detailed results are provided in Appendix~\ref{app:robust}.

\subsection{Self-InfCLIP for Training-Set Analysis}
We further study \textsc{InfCLIP} in a setting where the \emph{training dataset itself} is used as the target. This enables analysis of interactions among training samples and provides insight into how individual data points influence one another during CLIP training.

\begin{wrapfigure}{r}{0.58\linewidth}
    \centering

    \begin{subfigure}[b]{0.48\linewidth}
        \centering
        \includegraphics[width=\linewidth]{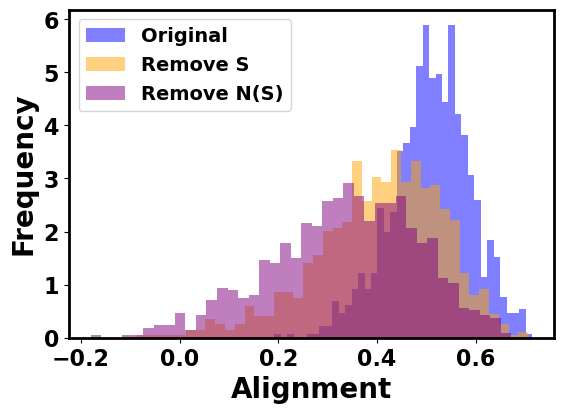}
        \caption{Alignment histogram}
        \label{fig:alignment_histogram}
    \end{subfigure}
    \hfill
    \begin{subfigure}[b]{0.48\linewidth}
        \centering
        \includegraphics[width=\linewidth]{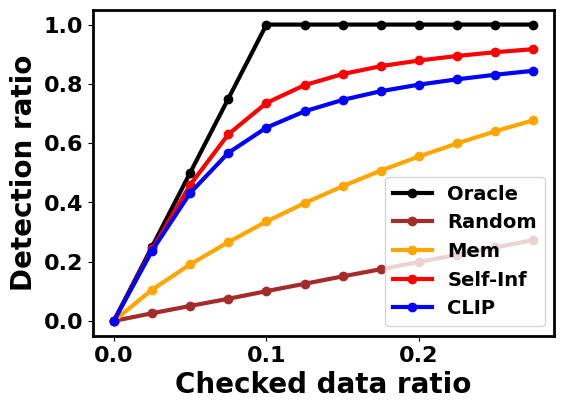}
        \caption{Noise detection}
        \label{fig:Noise_detection}
    \end{subfigure}

    \caption{\small Analysis of influential data.
    (a) Histogram of alignments under the three models.
    (b) Fraction of detected noisy samples among the inspected samples.}
    \label{fig:alignment_noise}
    \vspace{-5mm}
\end{wrapfigure}

\paragraph{Self-InfCLIP}
We first investigate how interactions among training samples affect image--text alignment.
To this end, we randomly select 1,000 image--text pairs $S$ from the CC3M training dataset $\mathcal{D}$.
For each sample $i \in S$, we identify its top-$k$ most influential neighbors according to the pairwise \textsc{InfCLIP} score $I(i,j)$.
We denote the union of these neighbors by
$N(S) = \bigcup_{i \in S} \argmax_{J \subseteq \mathcal{D},\, |J|=k} \sum_{j \in J} I(i,j)$.
We then train three CLIP models: (i) on the full dataset, (ii) with $S$ removed, and (iii) with $N(S)$ removed.
We then evaluate the alignment of samples in $S$ under the three models. As shown in Fig.~\ref{fig:alignment_histogram}, which reports histograms of alignment scores $\{f_i ^\top g_i\}_{i\in S}$ for the three settings, removing $N(S)$ leads to a substantially larger degradation in alignment on $S$ than removing $S$ itself. This indicates that alignment is not an intrinsic property of individual image-text pairs, but instead emerges from interactions among training samples. This observation motivates the need for a score that explicitly captures the strength of a sample's interactions with its most influential neighbors.

Accordingly, we define the \textsc{Self-InfCLIP} score of a sample $i$ as its aggregated influence on other top-$k$ most influential training samples, measuring its role within the learned multimodal structure:
\begin{equation}\label{eqn:SelfInf}
I_{self}(i,S) := \max_{\substack{J \subset S,\, |J| = k}} \sum_{j \in J} I(i,j),
\end{equation}
where $I(i,j)$ denotes the pairwise \textsc{InfCLIP} score and $k$ is the number of considered neighbors (set to $k=10$ in all experiments). This score reflects how strongly a sample interacts with its most related neighbors in the training set.

\paragraph{Connection to memorization score}
The interactions captured by \textsc{Self-InfCLIP} may also reflect how strongly a sample is supported by other training examples.
Samples with strong neighborhood interactions may be less sensitive to their own removal, as related samples can provide overlapping learning signals.
This suggests a connection between \textsc{Self-InfCLIP} and memorization, which measures how strongly a model relies on individual training samples.
To investigate this relationship, we compare \textsc{Self-InfCLIP} with the CLIPMem score \cite{CLIPMem}, which quantifies memorization through the change in alignment when a sample is removed:
$\mathrm{CLIPMem}(i) = f_T(x_i)^\top g_T(y_i) - f_{T \setminus i}(x_i)^\top g_{T \setminus i}(y_i)$,
where $(f_T, g_T)$ and $(f_{T \setminus i}, g_{T \setminus i})$ denote the encoders trained with and without sample $i$, respectively.
In practice, we approximate CLIPMem using ten leave-one-group-out models, as detailed in Appendix \ref{sec:exp_details}.

Empirically, we observe a Spearman rank correlation of $-0.60$ between \textsc{Self-InfCLIP} and CLIPMem, suggesting that samples strongly supported by their neighbors tend to exhibit lower memorization.
Computing CLIPMem is computationally expensive, as it requires repeatedly training models on different subsets of the dataset and estimating sample memorization from the resulting changes in image-text alignment.
In contrast, \textsc{Self-InfCLIP} directly measures interactions among training samples using only pretrained image and text features, without additional model training or gradient computation.
The observed negative correlation suggests that \textsc{Self-InfCLIP} can efficiently capture meaningful relationships among training samples that are associated with memorization.

\paragraph{Noise detection}

Leveraging this neighborhood-aware characterization, we apply \textsc{Self-InfCLIP} to noise detection.
We construct a noisy CC3M dataset by randomly selecting 10\% of the samples and shuffling their image-text pairs, thereby breaking semantic alignment while preserving marginal distributions.
Since a low \textsc{Self-InfCLIP} score indicates weak interaction with the local neighborhood, we hypothesize that such samples are likely to be noisy.
Accordingly, we identify noisy pairs by ranking samples in ascending order of \textsc{Self-InfCLIP}.
For each inspection ratio, we select the lowest-scoring samples and measure the fraction of artificially corrupted pairs successfully detected.
Figure~\ref{fig:alignment_noise} plots the fraction of noisy pairs detected as the inspected data ratio increases, and compares \textsc{Self-InfCLIP} with CLIPMem and the CLIP score. Detection performance can be evaluated by the area under the curve. \textsc{Self-InfCLIP} consistently outperforms both baselines in identifying noisy samples.

%% file: sections/5_Conclusion.tex
We proposed \textsc{InfCLIP}, an influence-inspired data valuation method for CLIP-style multimodal pretraining.
Unlike existing methods that heuristically combine alignment, target relevance, and uniqueness, \textsc{InfCLIP} unifies these criteria through a feature-based score derived from softmax-score gradient interactions and an NTK-inspired approximation.
This enables scalable data valuation using only pretrained features, without explicit gradient computation or retraining.
Its positive correlations with existing influence and attribution measures further support the influence-inspired formulation.
Extensive subset-selection experiments demonstrate its effectiveness across diverse pretraining datasets and downstream benchmarks, while \textsc{Self-InfCLIP} enables efficient noise detection and memorization-aware training-set analysis.
Overall, \textsc{InfCLIP} provides a unified and scalable framework for effective data selection and training-set analysis in multimodal pretraining.
Broader impacts and limitations are discussed in Appendix \ref{app:limitations}.

%% file: appendix/limitations.tex
\section{Broader Impacts and Limitations}\label{app:limitations}

\paragraph{Broader Impacts}
This work presents a principled and efficient approach to data valuation for large-scale multimodal datasets.
By moving beyond heuristic combinations of alignment-based scores, our method provides a more systematic way to assess the usefulness of image-text pairs for representation learning.
The proposed approach enables scalable data selection and dataset analysis without additional model retraining, making it practical for web-scale settings.
As a result, it can help reduce training costs and support the development of more reliable and effective multimodal models.
However, because \textsc{InfCLIP} relies on pretrained representations and target distributions, it may inherit or amplify existing biases and may undervalue underrepresented domains.
Practical use should therefore include bias, fairness, and domain-coverage analyses.

\paragraph{Limitations}
Our evaluation is limited to CLIP pretraining on CC3M, CC12M, and DataComp-small with standard CLIP architectures, and scaling the analysis to larger web-scale datasets, larger backbones, and longer training schedules remains future work.
Because \textsc{InfCLIP} uses representations from a pretrained teacher model, it may inherit teacher-specific biases and may undervalue samples from domains that are poorly represented by the teacher.
Moreover, \textsc{InfCLIP} is target-dependent: the selected subset is optimized for the specified target distribution, which may not always align with all downstream applications.

%% file: appendix/infclip_score.tex
\section{Details of InfCLIP Score}

\subsection{Full derivation of the score}\label{sec:full_derivation}
In this section, we provide a more detailed derivation of the InfCLIP score.
The InfCLIP score decomposes into eight terms, each consisting of
(i) alignment-related softmax weights and
(ii) inner products between adjusted feature representations.
Under the NTK approximation, the image-encoder and text-encoder terms coincide, so the eight terms combine into the four terms of equation \ref{eq:infclip_score}.
In the main text, we present only a representative term
(e.g., $p_{ii} p_{tt} \bar{g}_i^\top \bar{g}_t$) for clarity.
Here, we provide the complete derivation.

\paragraph{Preliminaries.}
Let $f_i = F(x_i)$ and $g_i = G(y_i)$,
and define the symmetric softmax score $s_i = p_{ii} + q_{ii}$,
where $p_{ij} = \dfrac{\exp(f_i^\top g_j/\tau)}{\sum_k \exp(f_i^\top g_k/\tau)}$
and $q_{ij} = \dfrac{\exp(f_i^\top g_j/\tau)}{\sum_k \exp(f_k^\top g_j/\tau)}$.

For a generic softmax variable $\hat{z}_i = \dfrac{\exp(z_i)}{\sum_k \exp(z_k)}$,
the derivatives satisfy $\dfrac{\partial \hat{z}_i}{\partial z_i} = \hat{z}_i (1 - \hat{z}_i)$
and $\dfrac{\partial \hat{z}_i}{\partial z_j} = - \hat{z}_i \hat{z}_j$ for $j \neq i$.

We denote the Jacobian of the image encoder output with respect to its parameters as $J_i := \nabla_F f_i \in \mathbb{R}^{d \times |F|}$.
Similarly, we denote the Jacobian of the text encoder as $K_i$.

\paragraph{Gradient of $p_{ii}$.}
Since $p_{ii}$ depends only on the image feature $f_i$, we have
$\partial p_{ii}/\partial f_k = 0$ for $k \neq i$, and thus
$$
\frac{\partial p_{ii}}{\partial F}
=\sum_k \frac{\partial f_k}{\partial F}\frac{\partial p_{ii}}{\partial f_k}
= J_i^\top \frac{\partial p_{ii}}{\partial f_i}.
$$
Let $z_k = f_i^\top g_k / \tau$, so that $p_{ik} = \hat{z}_k$.
Using the softmax derivatives and the chain rule, we obtain
$$
\frac{\partial p_{ii}}{\partial f_i}
=
\frac{1}{\tau}
\sum_k
\frac{\partial (f_i^\top g_k)}{\partial f_i}
\frac{\partial \hat{z}_i}{\partial z_k}
=
\frac{1}{\tau}
\sum_k
g_k
\frac{\partial \hat{z}_i}{\partial z_k}.
$$
The softmax derivative gives
$$
\frac{\partial \hat{z}_i}{\partial z_k}
=
\begin{cases}
\hat{z}_i(1-\hat{z}_i), & k=i,\\
-\hat{z}_i\hat{z}_k, & k\neq i,
\end{cases}
$$
and therefore
$$
\frac{\partial p_{ii}}{\partial f_i}
=
\frac{1}{\tau}
\left(
p_{ii}(1-p_{ii}) g_i
-
\sum_{k\neq i} p_{ii}p_{ik} g_k
\right).
$$
Rearranging terms yields
$$
\frac{\partial p_{ii}}{\partial f_i}
=
\frac{1}{\tau} p_{ii}
\left(
g_i - \sum_k p_{ik} g_k
\right)
=
\frac{1}{\tau} p_{ii}\,\bar{g}_i,
$$
where $\bar{g}_i := g_i - \sum_k p_{ik} g_k$.
Therefore,
$$
\frac{\partial p_{ii}}{\partial F}
=
\frac{1}{\tau} p_{ii} J_i^\top \bar{g}_i.
$$

\paragraph{Gradient of $q_{ii}$.}
Since $q_{ii}$ depends on all image features $\{f_k\}$, we have
$$
\frac{\partial q_{ii}}{\partial F}
=\sum\limits_k \frac{\partial f_k}{\partial F}\frac{\partial q_{ii}}{\partial f_k}
=\sum\limits_k J_k^\top \frac{\partial q_{ii}}{\partial f_k}.
$$
Substitute $f_k^\top g_i/\tau$ as $z_k$, so that $q_{ki}=\hat{z}_k$.
Using the softmax derivatives and chain rules, we obtain
$$
\frac{\partial q_{ii}}{\partial f_k}
=
\frac{1}{\tau}
\frac{\partial (f_k^\top g_i)}{\partial f_k}
\frac{\partial \hat{z}_i}{\partial z_k}
=
\frac{1}{\tau} g_i
\begin{cases}
\hat{z}_i(1-\hat{z}_i), & k=i,\\
-\hat{z}_i\hat{z}_k, & k\neq i,
\end{cases}
$$
i.e.,
$$
\frac{\partial q_{ii}}{\partial f_i}
=
\frac{1}{\tau}q_{ii}(1-q_{ii})g_i,
\qquad
\frac{\partial q_{ii}}{\partial f_k}
=
-\frac{1}{\tau}q_{ii}q_{ki}g_i\quad(k\neq i).
$$
Therefore,
$$
\frac{\partial q_{ii}}{\partial F}
=
\sum\limits_k J_k^\top \frac{\partial q_{ii}}{\partial f_k}
=
\frac{1}{\tau}q_{ii}\left(J_i^\top-\sum\limits_k q_{ki}J_k^\top\right)g_i
=
\frac{1}{\tau}q_{ii}\bar{J}_i^\top g_i,
$$
where $\bar{J}_i := J_i-\sum_k q_{ki}J_k$.

Therefore, we obtain
$$
\frac{\partial s_i}{\partial F}
=\frac{1}{\tau} \left( p_{ii} J_i^\top \bar{g}_i + q_{ii} \bar{J}_i^\top g_i \right),
$$

Symmetrically, for the gradient with respect to $G$, we obtain 
$$
\frac{\partial s_i}{\partial G}
=\frac{1}{\tau}\left(p_{ii} \bar{K}_i^\top f_i + q_{ii} K_i^\top \bar{f}_i \right),
$$

where $J_i := \nabla_F f_i$, $K_i := \nabla_G g_i$,
$\bar{g}_i := g_i - \sum_k p_{ik} g_k$,
$\bar{f}_i := f_i - \sum_k q_{ki} f_k$,
$\bar{J}_i := J_i - \sum_k q_{ki} J_k$,
and $\bar{K}_i := K_i - \sum_k p_{ik} K_k$.
The influence between samples $i$ and $t$ is
$I(i,t) = \nabla_F s_i \cdot \nabla_F s_t + \nabla_G s_i \cdot \nabla_G s_t$.

\paragraph{Inner product of gradients.}
The inner product over the image/text encoder parameters expands into four terms:

\begin{align}
&\nabla_F s_i \cdot \nabla_F s_t \nonumber \\
&\quad= \frac{1}{\tau^2}
\Big[
p_{ii} p_{tt}\, \bar{g}_i^\top ( J_i J_t^\top ) \bar{g}_t
+ p_{ii} q_{tt}\, \bar{g}_i^\top ( J_i \bar{J}_t^\top ) g_t
+ q_{ii} p_{tt}\, g_i^\top ( \bar{J}_i J_t^\top ) \bar{g}_t
+ q_{ii} q_{tt}\, g_i^\top ( \bar{J}_i \bar{J}_t^\top ) g_t
\Big], \nonumber \\
&\nabla_G s_i \cdot \nabla_G s_t \nonumber \\
&\quad= \frac{1}{\tau^2}
\Big[
p_{ii} p_{tt}\, f_i^\top (\bar{K}_i \bar{K}_t^\top) f_t
+ p_{ii} q_{tt}\, f_i^\top (\bar{K}_i K_t^\top) \bar{f}_t
+ q_{ii} p_{tt}\, \bar{f}_i^\top (K_i \bar{K}_t^\top) f_t
+ q_{ii} q_{tt}\, \bar{f}_i^\top (K_i K_t^\top) \bar{f}_t
\Big]. \label{eq:full_definition_ntk}
\end{align}

To obtain a closed-form feature expression, we apply a kernel-style
approximation that maps Jacobian interactions to (centered) feature
similarities, consistent with the centered structure of $\bar{J}_i$.
Specifically, we use
$J_i J_t^\top \approx (f_i^\top f_t) I_d$,
$J_i \bar{J}_t^\top \approx (f_i^\top \bar{f}_t) I_d$,
$\bar{J}_i J_t^\top \approx (\bar{f}_i^\top f_t) I_d$, and
$\bar{J}_i \bar{J}_t^\top \approx (\bar{f}_i^\top \bar{f}_t) I_d$.
Similarly, approximation of text encoder Jacobian is
$K_i K_t^\top \approx (g_i^\top g_t) I_d$,
$K_i \bar{K}_t^\top \approx (g_i^\top \bar{g}_t) I_d$,
$\bar{K}_i K_t^\top \approx (\bar{g}_i^\top g_t) I_d$, and
$\bar{K}_i \bar{K}_t^\top \approx (\bar{g}_i^\top \bar{g}_t) I_d$.

Substituting these approximations yields
\begin{align}
&\nabla_F s_i \cdot \nabla_F s_t \nonumber \\
&\quad\approx \frac{1}{\tau^2}
\Big[
p_{ii}p_{tt}(f_i^\top f_t)(\bar{g}_i^\top \bar{g}_t)
+
p_{ii}q_{tt}(f_i^\top \bar{f}_t)(\bar{g}_i^\top g_t)
+
q_{ii}p_{tt}(\bar{f}_i^\top f_t)(g_i^\top \bar{g}_t)
+
q_{ii}q_{tt}(\bar{f}_i^\top \bar{f}_t)(g_i^\top g_t)
\Big], \nonumber \\
&\nabla_G s_i \cdot \nabla_G s_t \nonumber \\
&\quad\approx \frac{1}{\tau^2}
\Big[
p_{ii}p_{tt}(\bar{g}_i^\top \bar{g}_t)(f_i^\top f_t)
+
p_{ii}q_{tt}(\bar{g}_i^\top g_t)(f_i^\top \bar{f}_t)
+
q_{ii}p_{tt}(g_i^\top \bar{g}_t)(\bar{f}_i^\top f_t)
+
q_{ii}q_{tt}(g_i^\top g_t)(\bar{f}_i^\top \bar{f}_t)
\Big]. \label{eq:full_definition_feature}
\end{align}

\paragraph{Final expression.}
Combining the two encoder contributions, the influence score is approximated by four feature-product terms:
\begin{equation*}
\begin{split}
&I(i,t) = \nabla_F s_i \cdot \nabla_F s_t
 + \nabla_G s_i \cdot \nabla_G s_t\\
&\quad\approx \frac{2}{\tau^2}
\Big[
p_{ii}p_{tt}(f_i^\top f_t)(\bar{g}_i^\top \bar{g}_t)
+
p_{ii}q_{tt}(f_i^\top \bar{f}_t)(\bar{g}_i^\top g_t)
+
q_{ii}p_{tt}(\bar{f}_i^\top f_t)(g_i^\top \bar{g}_t)
+
q_{ii}q_{tt}(\bar{f}_i^\top \bar{f}_t)(g_i^\top g_t)
\Big].
\end{split}
\end{equation*}

\subsection{Last Layer Approximation}\label{sec:lastlayer}
Many works adopt calculating gradients only at the last layer, rather than obtaining the true gradients of the full model, and utilize them as a proxy for the full gradient \cite{TracIn, EL2N}.
When we consider only the gradients of the last layer of the CLIP model, which is a linear projection layer, quite similar results can be obtained under appropriate assumptions.

Let $x_i, y_i$ denote the outputs of the image and text encoders, respectively.
They are passed through linear projection layers $F, G$ to produce the final features $f_i, g_i$.
We define the model parameters as $\theta = \{F, G\}$, with
$x \in \mathbb{R}^{d_x}$ denoting image features,
$f \in \mathbb{R}^{d}$ denoting projected image features,
$y \in \mathbb{R}^{d_y}$ denoting text features,
and $g \in \mathbb{R}^{d}$ denoting projected text features.

The projection layers are defined as
$F \in \mathbb{R}^{d \times d_x}$ for the image encoder and
$G \in \mathbb{R}^{d \times d_y}$ for the text encoder,
with image encoding $f = Fx$ and text encoding $g = Gy$.

Since $F$ and $G$ are linear maps, their Jacobians take simple forms.
In particular,
\[
\frac{\partial f_i}{\partial F} = x_i^\top \otimes I_d,
\qquad
\frac{\partial g_i}{\partial G} = y_i^\top \otimes I_d.
\]

Therefore, the Jacobian matrix product of image features satisfies
\[
J_iJ_t^\top=(x_i^\top\otimes I_d)(x_t\otimes I_d)=(x_i^\top x_t)I_d,
\]
where $I_d$ is the $d\times d$ identity matrix and $\otimes$ denotes the Kronecker product.

As a result, the inner product between gradients of alignment scores with respect to the image projection layer can be written as
\begin{equation*}
\begin{split}
&\nabla_F s_i \cdot \nabla_F s_t \\
&\quad= \frac{1}{\tau^2}
\Big[
p_{ii} p_{tt}\, (\bar{g}_i^\top \bar{g}_t)(x_i^\top x_t)
+ p_{ii} q_{tt}\, (\bar{g}_i^\top g_t)(x_i^\top \bar{x}_t)
+ q_{ii} p_{tt}\, (g_i^\top \bar{g}_t)(\bar{x}_i^\top x_t)
+ q_{ii} q_{tt}\, (g_i^\top g_t)(\bar{x}_i^\top \bar{x}_t)
\Big],
\end{split}
\end{equation*}
where $\bar{x}_i := x_i - \sum_k q_{ki} x_k$ is the centered image feature induced by the text-side softmax.

\begin{figure}
    \centering
    \includegraphics[width=0.5\linewidth]{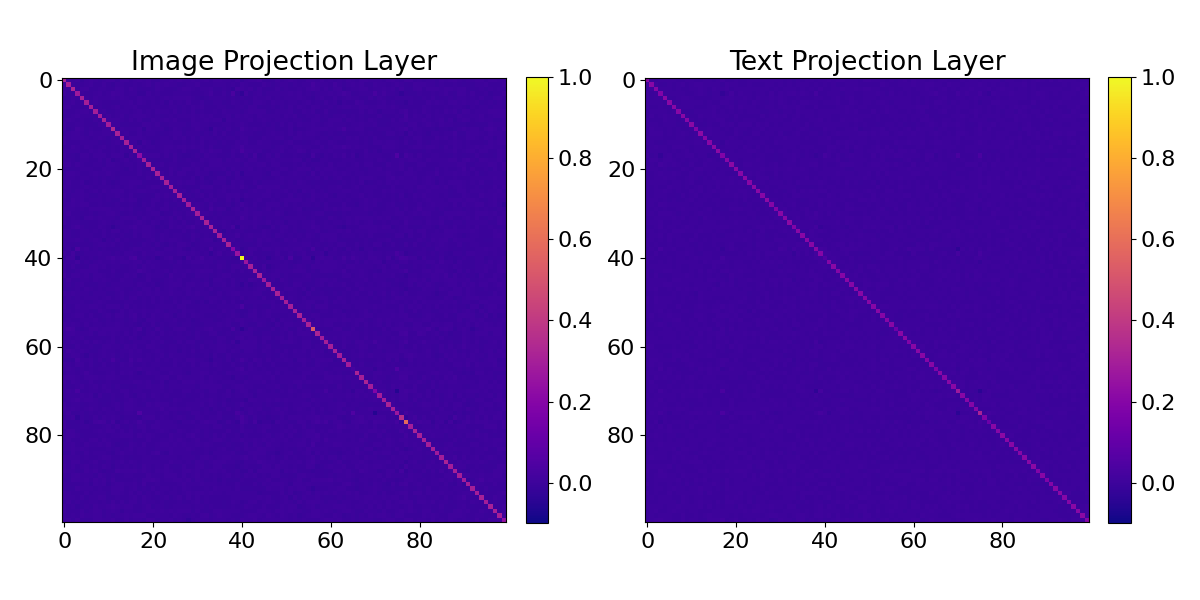}
    \caption{Heatmap of $F^\top F$ and $G^\top G$ for 100 $\times$ 100 indices. They are well approximated by diagonal matrices.}
    \label{fig:projection_layers}
\end{figure}
Here we utilize the empirical observation that the last projection layers $F$ and $G$
are close to orthogonal, which implies
\[
f_i^\top f_t = x_i^\top F^\top F x_t \simeq c_1\,x_i^\top x_t,
\qquad
g_i^\top g_t = y_i^\top G^\top G y_t \simeq c_2\,y_i^\top y_t.
\]

We explicitly compute $F^\top F$ and $G^\top G$ and observe that they are well
approximated by a constant multiple of the identity matrix.
This validates the approximation that inner products in the input feature space
are preserved after projection.
Figure~\ref{fig:projection_layers} visualizes this behavior by showing colormap
representations of $F^\top F$ and $G^\top G$, where the diagonal entries are
dominant and nearly constant, while the off-diagonal entries remain small.

Therefore, under the approximate orthogonality of the last projection layers, the gradient interactions can be approximated as

\begin{equation*}
\begin{split}
&\nabla_F s_i \cdot \nabla_F s_t \\
&\quad\simeq \frac{1}{\tau^2 c_1}
\Big[
p_{ii}p_{tt}\, (f_i^\top f_t)(\bar{g}_i^\top \bar{g}_t)
+ p_{ii}q_{tt}\, (f_i^\top \bar{f}_t)(\bar{g}_i^\top g_t)
+ q_{ii}p_{tt}\, (\bar{f}_i^\top f_t)(g_i^\top \bar{g}_t)
+ q_{ii}q_{tt}\, (\bar{f}_i^\top \bar{f}_t)(g_i^\top g_t)
\Big].
\end{split}
\end{equation*}
Similarly, for the text projection layer,
\begin{equation*}
\begin{split}
&\nabla_G s_i \cdot \nabla_G s_t \\
&\quad\simeq \frac{1}{\tau^2 c_2}
\Big[
p_{ii}p_{tt}\, (\bar{g}_i^\top \bar{g}_t)(f_i^\top f_t)
+ p_{ii}q_{tt}\, (\bar{g}_i^\top g_t)(f_i^\top \bar{f}_t)
+ q_{ii}p_{tt}\, (g_i^\top \bar{g}_t)(\bar{f}_i^\top f_t)
+ q_{ii}q_{tt}\, (g_i^\top g_t)(\bar{f}_i^\top \bar{f}_t)
\Big].
\end{split}
\end{equation*}

%% file: appendix/ablation_calculation.tex
\section{Ablation Experiment of InfCLIP Score}\label{sec:Ablation}

\begin{figure*}
    \centering
    \includegraphics[width=0.99\linewidth]{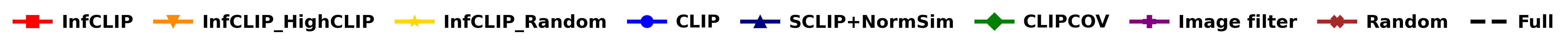} \\
    \includegraphics[width=0.32\linewidth]{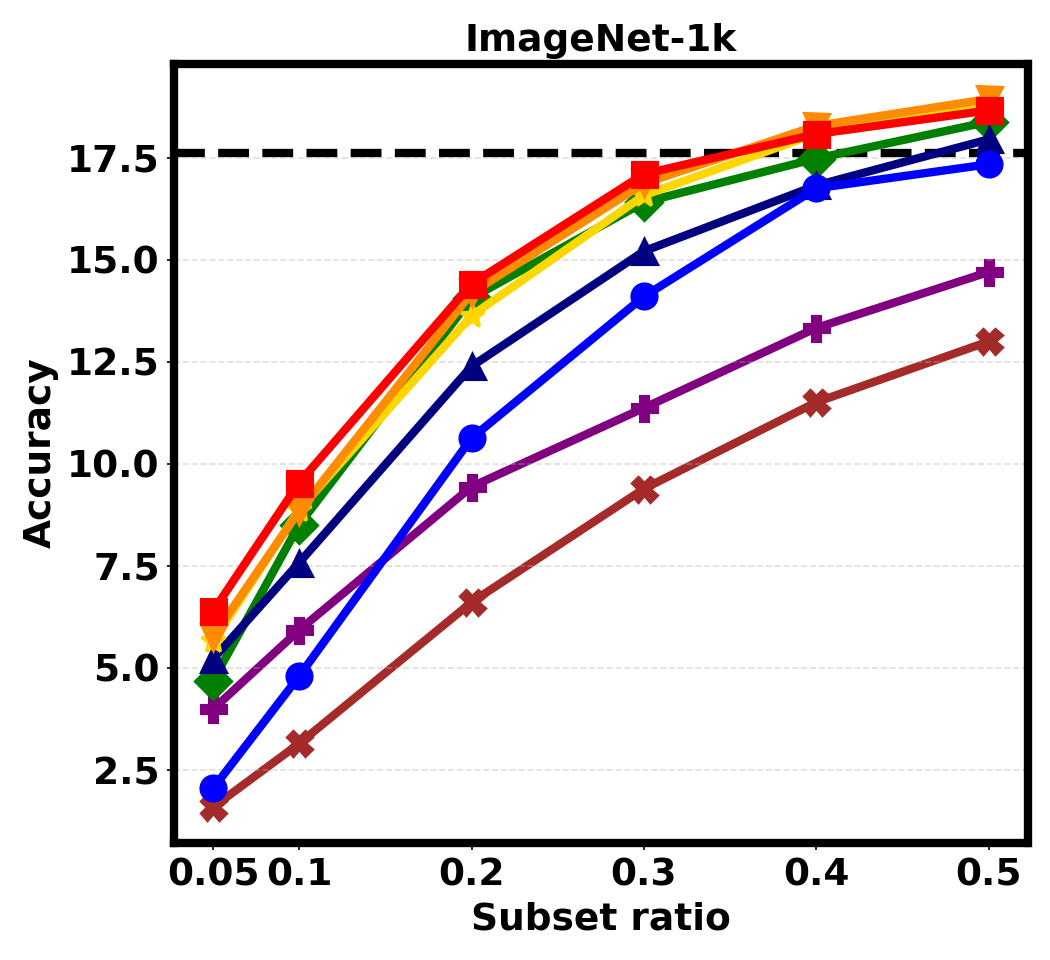}
    \includegraphics[width=0.32\linewidth]{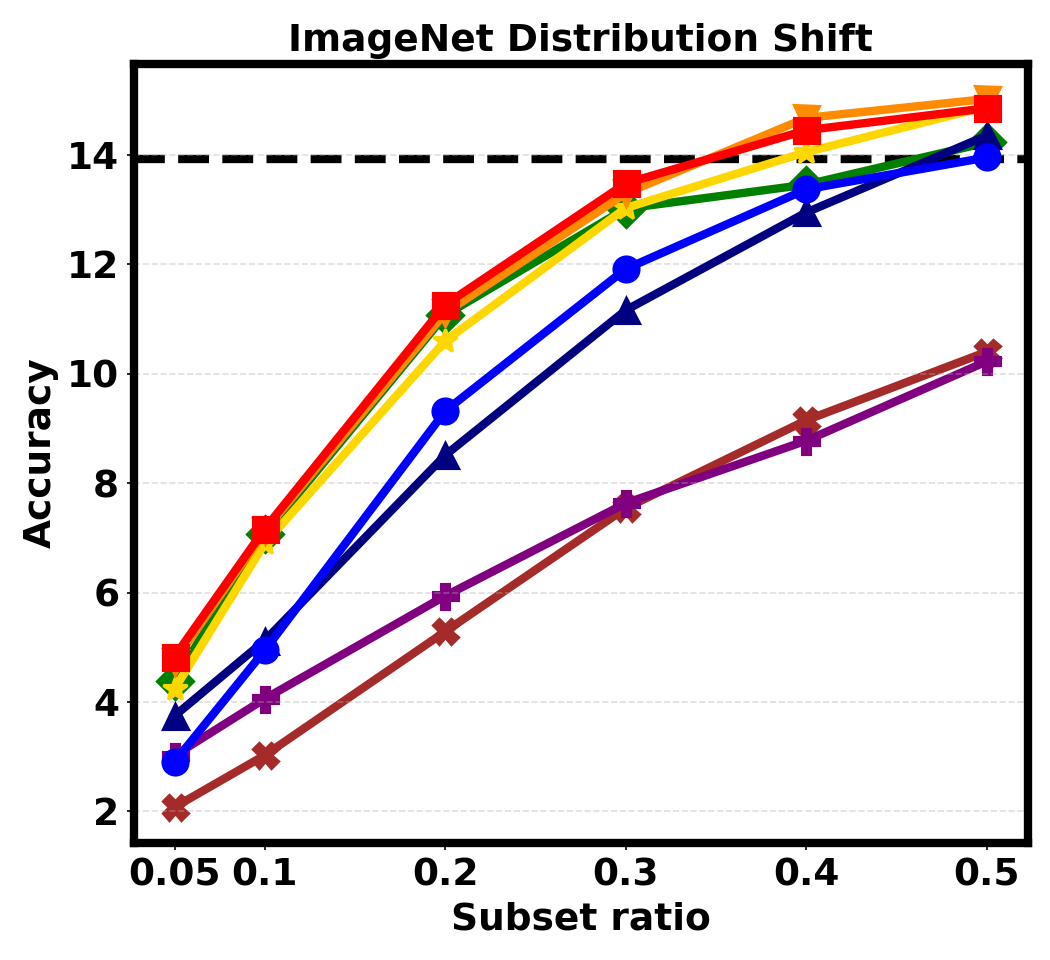}
    \includegraphics[width=0.32\linewidth]{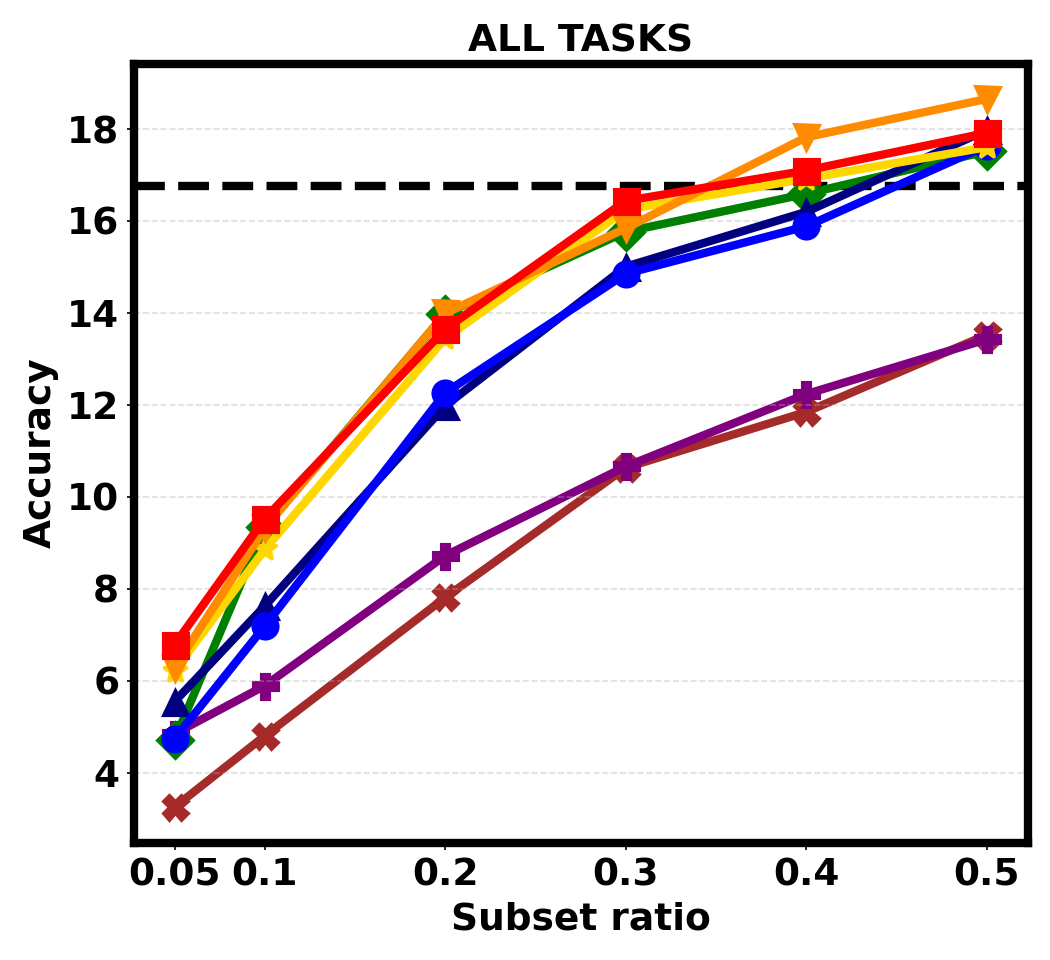}
    \caption{Prototype ablation for the target set $T$. 
    We vary the prototype construction (class-mean, high-CLIP, random) and report zero-shot performance of CLIP models trained on CC3M subsets. 
    The overall trends and the relative advantage of InfCLIP remain consistent across prototype choices.}
    \label{fig:ablation_prototype}
\end{figure*}

\paragraph{Prototype selection.}
In Section \ref{sec:SubsetSelection}, we compute the target-aware score as $I(i,T)=\max_{t\in T} I(i,t)$ using a prototype target set $T$ instead of the full ImageNet dataset. 
Here, we ablate the construction of $T$ and verify that InfCLIP is insensitive to reasonable prototype choices. 
In the main experiments, we construct class-mean prototypes using ImageNet class labels. 
For each class $c\in\mathcal{C}$, we form the image prototype by averaging image embeddings, and form the text prototype by averaging prompt-encoded class-name embeddings using the standard CLIP prompt templates \cite{CLIP}. 
Concretely, letting $\mathcal{I}_c$ be ImageNet images of class $c$ and $\mathcal{P}$ be a set of prompts, we define
\begin{equation*}
\mu^f_c=\frac{1}{|\mathcal{I}_c|}\sum_{x\in\mathcal{I}_c} f(x),\quad 
\mu^g_c=\frac{1}{|\mathcal{P}|}\sum_{p\in\mathcal{P}} g(p(c)).
\end{equation*}
We then set $T=\{(\mu^f_c,\mu^g_c)\}_{c\in\mathcal{C}}$, yielding $|T|=|\mathcal{C}|=1000$ prototypes for ImageNet. 
We compare this class-mean construction against two alternatives. 
(i) \textbf{High-CLIP prototype} selects one representative per class by choosing the image whose alignment with its class prompt is the highest under the teacher CLIP features. 
(ii) \textbf{Random prototype} selects one random image per class and pairs it with the corresponding prompt-averaged text embedding. 
Figure \ref{fig:ablation_prototype} shows that these variants produce similar subset-selection performance across ImageNet, ImageNet-shift benchmarks, and the aggregated task suite. 
This indicates that the gains of InfCLIP do not rely on a particular prototype construction, and that compact targets provide a reliable surrogate for the full target dataset. 

\paragraph{Softmax parameters for score computation.}
We ablate the parameters used when computing the softmax probabilities $p_{ij}$ and $q_{ij}$ (and hence $\bar{g}_i$ and $\bar{f}_i$) for InfCLIP scoring, focusing on the scoring temperature $\tau$ (equivalently, the logit scale $1/\tau$), which controls the sharpness of the softmax distribution and the magnitude of the adjusted feature vectors.
As discussed in the main text, pretrained CLIP models often induce overly sharp softmax distributions.
In this regime, $p_{ii} \approx 1$ and $p_{ik} \approx 0$ for $k \neq i$, causing the expectation $\sum_k p_{ik} g_k$ to collapse to $g_i$ and driving the centered vector $\bar{g}_i = g_i - \sum_k p_{ik} g_k$ toward zero.
This effect is amplified for large logit scales (small $\tau$), leading to vanishing gradients and weak influence signals.
Temperature rescaling mitigates this issue by softening the softmax distribution, allowing multiple samples to contribute to the expectation.
This can be interpreted as approximating a less-confident regime, similar to earlier or intermediate stages of training, where per-sample interactions are more distinguishable.

\begin{figure}[h!]
    \centering
    \begin{minipage}[t]{0.55\textwidth}
        \vspace{0pt}
        We vary the scoring temperature $\tau$ and evaluate the stability of the induced sample ranking.
        Figure~\ref{fig:ablation_softmax} reports rank correlations between InfCLIP scores computed with different logit scales.
        The ranking remains stable over a broad, moderate range of logit scales (e.g., 5 $\sim$ 20 in our setting), indicating robustness to this choice.
        Extremely small logit scales over-flatten the distribution and reduce discriminability, while extremely large logit scales make $p_{ij}$ and $q_{ij}$ overly sharp, again driving $\bar{g}_i$ and $\bar{f}_i$ toward zero.
        
        Temperature rescaling is applied only during score computation and does not modify the pretrained CLIP model.
        In all main experiments, we therefore use moderate temperatures, specifically $\tau=0.1$ (logit scale $=10$) for target-dataset InfCLIP and $\tau=1$ (logit scale $=1$) for Self-InfCLIP.
        We also find that beyond moderate batch sizes, the ranking and downstream subset-selection trends remain unchanged.
    \end{minipage}
    \hfill
    \begin{minipage}[t]{0.40\textwidth}
        \vspace{0pt}
        \centering
        \includegraphics[width=\linewidth]{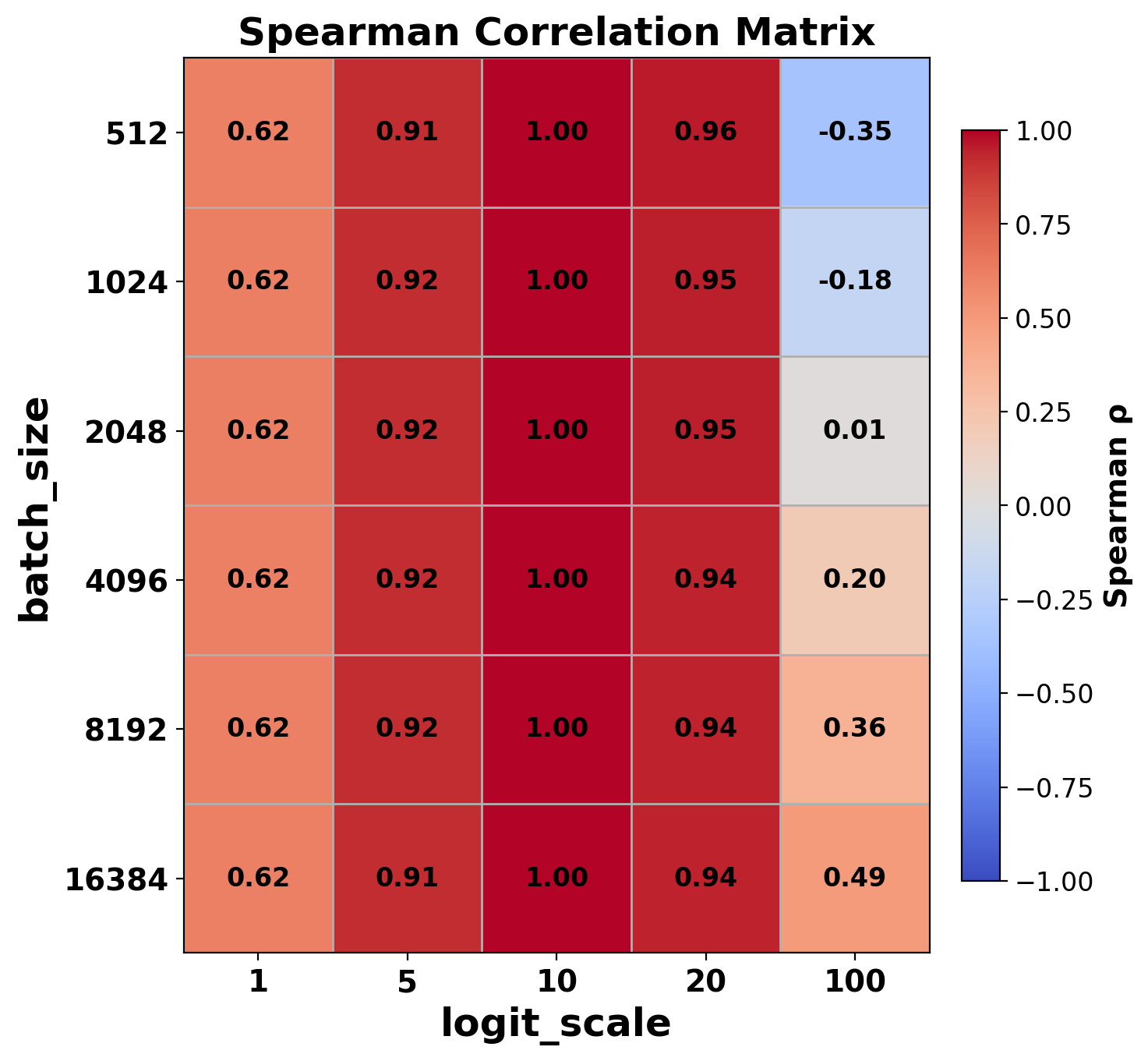}
        \caption{Rank correlations of InfCLIP scores across batch size and logit scale choices.}
        \label{fig:ablation_softmax}
    \end{minipage}
\end{figure}

%% file: appendix/robustness.tex
\section{Robustness across Training and Target Distributions}\label{app:robust}

\setcounter{topnumber}{3}
\setcounter{totalnumber}{4}
\renewcommand{\topfraction}{0.95}
\renewcommand{\textfraction}{0.05}

\paragraph{Subset selection on DataComp dataset}
We further conduct subset-selection experiments on DataComp-small
\cite{DataComp}, a noisier and more diverse web-scale image--text
dataset, following the protocol in Section~\ref{sec:SubsetSelection}.
We compute \textsc{InfCLIP} scores using ImageNet prototypes
as the target set and select the top-scoring 5\%, 10\%, and 20\%
of the training data.
We compare \textsc{InfCLIP} with CLIP score and SCLIP+NormSim,
and evaluate the trained models on ImageNet-1K, six ImageNet
distribution-shift benchmarks, and seven VTAB-Natural tasks.

The official DataComp-small candidate pool contains approximately
12.8 million image--text pairs, whereas the Hugging Face release
used in our experiments contains approximately 8.38 million
available pairs. All subset ratios and the full-data baseline
are defined relative to this available training pool.
We use a fixed training budget corresponding to one full-data
epoch, keeping the total number of optimization steps
identical across all subset ratios and methods.

\begin{table}[!htbp]
\centering
\caption{Subset-selection results on DataComp-small.
All Tasks denotes the average over 14 classification benchmarks.
The full-data baseline uses the same optimization-step budget
as the selected subsets.}
\label{tab:datacomp_small}
\small
\setlength{\tabcolsep}{5pt}
\begin{tabular}{llcccc}
\toprule
Subset ratio & Evaluation dataset & CLIP & SCLIP+NormSim
& \textsc{InfCLIP} & Full \\
\midrule
\multirow{3}{*}{5\%}
& ImageNet-1K       & 1.54 & 2.97 & \textbf{7.78} & 4.56 \\
& ImageNet Shift    & 2.50 & 3.07 & \textbf{7.07} & 5.23 \\
& All Tasks         & 10.50 & 11.74 & \textbf{15.19} & 14.63 \\
\midrule
\multirow{3}{*}{10\%}
& ImageNet-1K       & 3.70 & 5.73 & \textbf{10.82} & 4.56 \\
& ImageNet Shift    & 4.92 & 5.31 & \textbf{9.75} & 5.23 \\
& All Tasks         & 13.92 & 15.01 & \textbf{18.90} & 14.63 \\
\midrule
\multirow{3}{*}{20\%}
& ImageNet-1K       & 7.31 & 8.39 & \textbf{12.28} & 4.56 \\
& ImageNet Shift    & 7.69 & 7.94 & \textbf{11.28} & 5.23 \\
& All Tasks         & 17.57 & 17.71 & \textbf{21.06} & 14.63 \\
\bottomrule
\end{tabular}
\end{table}

As shown in Table~\ref{tab:datacomp_small},
\textsc{InfCLIP} consistently outperforms the baselines
across all subset ratios and evaluation benchmarks.
Notably, selecting only 5\% of the training data already
outperforms the full-data baseline, demonstrating its
effectiveness on a noisier and more diverse pretraining corpus.
We note that our training architecture and implementation
differ from the official DataComp protocol, and thus
we do not directly compare absolute performance with
the official DataComp leaderboard.

\paragraph{Choice of target dataset}
In the main experiments, we use ImageNet prototypes
as the target set for computing \textsc{InfCLIP} scores.
We further investigate how the choice of target dataset
affects downstream performance by incorporating
COCO and VOC2007 as additional target distributions.

\textbf{COCO retrieval.}
We consider three target configurations:
(i) ImageNet prototypes (1,000 classes),
(ii) ImageNet prototypes augmented with 100 randomly
selected COCO image-text pairs, and
(iii) 5,000 COCO image-text pairs alone.
For each configuration, we compute \textsc{InfCLIP}
scores on CC3M and train CLIP models from scratch
using the top-scoring 20\% or 30\% subsets.
We evaluate ImageNet-1K zero-shot accuracy and
COCO image-text retrieval performance using Mean Recall.

As shown in Table~\ref{tab:target_coco},
using COCO as the target dataset achieves the highest
retrieval performance but reduces ImageNet accuracy.
In contrast, augmenting ImageNet prototypes with only
100 COCO samples improves retrieval performance while
largely preserving ImageNet accuracy.
This demonstrates that \textsc{InfCLIP} can adapt
data selection to different downstream objectives
through target-set composition.

\textbf{VOC2007 classification.}
We additionally extend the ImageNet target set with
VOC2007 samples and evaluate the resulting models
on both ImageNet and VOC2007.
Table~\ref{tab:target_voc} compares performance
before and after incorporating VOC2007 into the
target set.

\begin{table}[!htbp]
\centering
\caption{Effect of target-set composition on ImageNet
classification and COCO image--text retrieval.
Models are trained on the selected subsets of CC3M.
Mean Recall averages Recall@1, Recall@5, and Recall@10
over both retrieval directions.}
\label{tab:target_coco}
\small
\setlength{\tabcolsep}{7pt}
\begin{tabular}{clcc}
\toprule
Subset ratio & Target dataset & ImageNet-1k & COCO Mean Recall \\
\midrule
\multirow{3}{*}{20\%}
& ImageNet        & \textbf{14.71} & 20.66 \\
& ImageNet + COCO & 14.50 & 22.59 \\
& COCO            & 10.82 & \textbf{23.25} \\
\midrule
\multirow{3}{*}{30\%}
& ImageNet        & \textbf{17.10} & 26.40 \\
& ImageNet + COCO & 17.00 & 27.88 \\
& COCO            & 13.66 & \textbf{28.57} \\
\bottomrule
\end{tabular}
\end{table}

\begin{table}[!htbp]
\centering
\caption{Effect of adding VOC2007 to the ImageNet
target set. We compare performance using ImageNet
targets alone and ImageNet targets augmented with VOC2007.}
\label{tab:target_voc}
\small
\setlength{\tabcolsep}{9pt}
\begin{tabular}{llcc}
\toprule
Subset & Evaluation dataset
& ImageNet target & ImageNet + VOC target \\
\midrule
\multirow{2}{*}{10\%}
& ImageNet-1K & 9.19 & 9.40 \\
& VOC2007    & 16.04 & \textbf{25.41} \\
\midrule
\multirow{2}{*}{30\%}
& ImageNet-1K & \textbf{17.10} & 17.04 \\
& VOC2007    & 23.28 & \textbf{30.51} \\
\bottomrule
\end{tabular}
\end{table}

As shown in Table~\ref{tab:target_voc},
incorporating VOC2007 into the target set substantially
improves VOC2007 performance while maintaining
comparable ImageNet accuracy.
Together with the COCO results, this demonstrates
that \textsc{InfCLIP} can flexibly incorporate
additional target distributions without substantially
sacrificing performance on the original target task.

\clearpage

%% file: appendix/experiment_details.tex
\section{Experiment Details}\label{sec:exp_details}
This section provides implementation details for all experiments, including (i) score computation for \textsc{InfCLIP}, (ii) training from scratch on selected subsets, and (iii) evaluation protocols.
Unless stated otherwise, we use CC3M as the pretraining corpus and ImageNet as the target dataset for subset selection.
To make the target set scalable for scoring, we use an ImageNet \emph{prototype target} instead of the full ImageNet dataset when computing $I(i,T)$.

\paragraph{Optimization.}
We train models using AdamW with batch size 1024, learning rate $5\times 10^{-4}$, weight decay $0.1$, and $\epsilon=10^{-8}$.
Unless stated otherwise, all CLIP models are trained for a fixed number of epochs (specified per experiment) under the same optimizer configuration to ensure comparability across selection methods.
For pruning/subset experiments, we match the training budget across methods by using the same number of optimization steps.
Concretely, we compute the number of iterations required to traverse the (method-specific) subset once (one epoch) and fix the total number of iterations across different selection methods and subset ratios.
This prevents confounding effects from unequal training steps when the subset size differs.

\paragraph{Two models for scoring and training.}
We use two CLIP models for different purposes.
First, for \emph{feature extraction and score computation}, we use an OpenCLIP model pretrained on DataComp-XL: ``open\_clip.create\_model\_and\_transforms('ViT-B-32', pretrained='datacomp\_xl\_s13b\_b90k') ''. \cite{CLIP, DataComp}
Second, for \emph{training CLIP models from scratch} in all subset-training experiments (window training, pruning, subset selection, noise detection), we use the OpenAI CLIP architecture with a modified ResNet-50 visual encoder and a Transformer-based text encoder, with embedding dimension $512$.
The configuration is summarized below:
We use an OpenAI CLIP-style model with an RN50 visual backbone.
The model uses an embedding dimension of 512 and an input image resolution of $224$.
The ResNet visual encoder has layers $[3,4,6,3]$ with width $64$.
The text encoder is a Transformer with width $512$, $8$ attention heads, and $6$ layers.
We use a context length of $77$ tokens and a vocabulary size of $49{,}408$.

\paragraph{Evaluation Tasks.}
We report zero-shot accuracy on the ImageNet validation set, and additionally evaluate robustness and transfer using:
\begin{itemize}
    \item \textbf{ImageNet distribution shifts:}
    ImageNet-A~\cite{ImageNet-A},
    ImageNet-R~\cite{ImageNet-R},
    ImageNet-V2~\cite{ImageNet-V2},
    ImageNet-Sketch~\cite{ImageNet-sketch},
    ImageNet-O,
    and ObjectNet~\cite{ObjectNet}.
    \item \textbf{VTAB-Natural:}
    CIFAR-100, Caltech101, DTD, Flowers102, Oxford-IIIT Pets, SUN397, and SVHN from VTAB~\cite{VTAB}.
\end{itemize}

\paragraph{Window subset selection}
For the window-training experiment (Fig.~\ref{fig:Fig1}), we sort CC3M samples by their CLIP scores and construct \emph{window subsets} with width $w\%$ and start point $s$.
Each window contains samples whose score ranks fall within $[s\%, (s+w)\%]$.
We train a CLIP model from scratch on each window subset and evaluate zero-shot top-5 accuracy on ImageNet.
The window is slid by $10\%$, with $s \in \{0\%, 10\%, 20\%, \ldots, 90\%\}$, and we consider window widths $w \in \{5\%, 10\%\}$.
As shown in Fig.~\ref{fig:Fig1}, CLIP score-based selection does not induce a monotonic trend in downstream performance.
In contrast, selecting data based on \textsc{InfCLIP} scores yields a clear monotonic improvement, highlighting the advantage of influence-based data valuation over alignment-based heuristics.

\paragraph{High-score pruning.}
For the reverse pruning experiment (Tab.~\ref{tab:reversepruning}), we remove the top $r\%$ of samples from CC3M dataset according to each scoring method and train a CLIP model on the remaining data, and check zero-shot top-1 accuracy on ImageNet dataset.
A sharp performance drop after removing high InfCLIP scoring samples indicates that the method correctly identifies training examples that are critical for generalization.
Scores are computed in the same manner as in the subset selection experiment, using $I(i,T) = \max_{t \in T} I(i,t)$.

\paragraph{Subset selection.}
For each subset ratio (5, 10, 20, 30, 40, 50 \%), we select the top-scoring CC3M samples according to each scoring method and train a CLIP model from scratch on the selected subset.
We compare against several baselines described in the main text, including CLIP-score ranking, image filtering, SCLIP+NormSim, and CLIPCOV.

\paragraph{Noise detection and memorization score.}
For the noise detection experiment, we sample $1{,}000{,}000$ examples from CC3M (indices 0--999{,}999) and treat them as the full dataset.
We create a noisy CC3M variant by randomly selecting $10\%$ of samples and shuffling their image-text pairs, thereby breaking semantic alignment while preserving the marginal distributions.
We compute scores for all samples and treat low-scoring examples as noisy.

To compute memorization scores, we construct ten leave-one-group-out subsets.
Specifically, the 1M dataset is partitioned into ten disjoint groups:
group 1: [0--99{,}999], group 2: [100{,}000--199{,}999], $\ldots$, group 10: [900{,}000--999{,}999].
Each subset is formed by removing one group from the full dataset, resulting in a dataset of size 0.9M.
For each subset, we train a CLIP model and evaluate the alignment of the held-out group using the corresponding model.
That is, the model trained on the full dataset excluding the $k$-th group is used to compute the alignment scores of samples in the $k$-th group.
The memorization score is defined as the difference between the alignment obtained by the model trained on the full dataset and that obtained by the model trained on the corresponding subset.

\paragraph{Computational costs.}
The total computational cost of \textsc{InfCLIP} decomposes into feature extraction and softmax-based score computation.
Scoring requires a single forward pass to extract image and text features, followed by batch-wise softmax computations to form $p_{ij}$ and $q_{ij}$ and the corresponding adjusted representations.
No gradients, Hessians, or model retraining are required during scoring.
As a result, \textsc{InfCLIP} is substantially more efficient than retraining-based data valuation methods (e.g., leave-one-out memorization measures), which require training multiple models.

The main computational bottleneck arises from the softmax normalization within each batch.
For a batch size $B$, computing $p_{ij}$ and $q_{ij}$ requires forming a $B \times B$ similarity matrix, resulting in time and memory complexity that scale quadratically with $B$.
While larger batch sizes enable more global normalization across samples, they significantly increase memory consumption and runtime.
In practice, we therefore use moderate batch sizes that balance computational efficiency and score stability.

Table~\ref{tab:time} reports the wall-clock time required to compute \textsc{InfCLIP} scores on CC3M for different batch sizes.
We observe that moderate batch sizes achieve a favorable trade-off between computational cost and stability, without requiring excessively large batches.
The timing experiment is for score computation only and uses a single NVIDIA RTX A6000 GPU, while subset training is performed with multiple GPUs.
Score computation completes within 30 seconds in all settings.

\begin{table}[]
    \centering
    \caption{Time (seconds) for computing \textsc{InfCLIP} scores on CC3M with different batch sizes.}
    \label{tab:time}
    \begin{tabular}{c|c|c|c|c|c|c}
    \toprule
        Batch size&  512 & 1024 & 2048 & 4096 & 8192 & 16384\\ \midrule
        Time (Sec)& 24 & 15 & 18 & 19 & 24 & 31  \\
        \bottomrule
    \end{tabular}
\end{table}

\paragraph{Algorithm.}
We provide the full procedure for computing InfCLIP influence scores based on the softmax NTK formulation in Algorithm~\ref{alg:infclip-softmax-ntk}.
Here, $\tau$ denotes the temperature parameter, and the corresponding
logit scale is given by $1/\tau$.

\begin{algorithm}[t]
\caption{InfCLIP Influence via Softmax NTK}
\label{alg:infclip-softmax-ntk}
\begin{algorithmic}[1]

\REQUIRE
Training features $\{(f_i, g_i)\}_{i=1}^N$, 
Target features $\{(f_v, g_v)\}_{v=1}^M$, 
temperature $\tau$, batch size $B$

\ENSURE
Influence scores $\mathrm{Inf}(i)$ for each training sample $i$

\STATE Normalize all features:
\[
f \leftarrow \frac{f}{\|f\|_2}, \quad
g \leftarrow \frac{g}{\|g\|_2}
\]

\STATE Initialize influence list $\mathcal{I} \leftarrow \emptyset$

\FOR{each training batch $\mathcal{B} \subset \{1,\dots,N\}$ of size $B$}

    \STATE Extract batch features $\{f_i, g_i\}_{i \in \mathcal{B}}$

    \STATE Compute temperature-scaled logits:
    \[
    L_{ij} = \frac{1}{\tau} f_i^\top g_j
    \]

    \STATE Row-wise and column-wise softmax:
    \[
    p_{ij} = \mathrm{softmax}_j(L_{ij}), \quad
    q_{ij} = \mathrm{softmax}_i(L_{ij})
    \]

    \STATE Diagonal terms:
    \[
    p_i = p_{ii}, \quad q_i = q_{ii}
    \]

    \STATE Training-side centered features:
    \[
    \bar{f}_i = f_i - \sum_k q_{k i} f_k, \quad
    \bar{g}_i = g_i - \sum_j p_{i j} g_j
    \]

    \STATE Target-side logits:
    \[
    L_{vv'} = \frac{1}{\tau} f_v^\top g_{v'}
    \]

    \STATE Target-side softmax:
    \[
    p_{vv'} = \mathrm{softmax}_{v'}(L_{vv'}), \quad
    q_{vv'} = \mathrm{softmax}_{v}(L_{vv'})
    \]

    \STATE Target-side diagonal terms:
    \[
    p_v = p_{vv}, \quad q_v = q_{vv}
    \]

    \STATE Target-side centered features:
    \[
    \bar{f}_v = f_v - \sum_{v'} q_{v' v} f_{v'}, \quad
    \bar{g}_v = g_v - \sum_{v'} p_{v v'} g_{v'}
    \]

    \STATE Compute weight matrices:
    \[
    \begin{aligned}
    w^{pp}_{iv} &= p_i p_v, &
    w^{pq}_{iv} &= p_i q_v, &
    w^{qp}_{iv} &= q_i p_v, &
    w^{qq}_{iv} &= q_i q_v
    \end{aligned}
    \]

    \STATE \textbf{F-part similarity:}
    \[
    \begin{aligned}
    S^{F}_{iv} =
    &\, w^{pp}_{iv} (f_i^\top f_v)(\bar{g}_i^\top \bar{g}_v)
    + w^{pq}_{iv} (f_i^\top \bar{f}_v)(\bar{g}_i^\top g_v)
    + w^{qp}_{iv} (\bar{f}_i^\top f_v)(g_i^\top \bar{g}_v)
    + w^{qq}_{iv} (\bar{f}_i^\top \bar{f}_v)(g_i^\top g_v)
    \end{aligned}
    \]

    \STATE \textbf{G-part similarity:}
    \[
    \begin{aligned}
    S_{iv}^{G}
    ={}&w_{iv}^{pp}(\bar{g}_i^\top\bar{g}_v)(f_i^\top f_v)
    +w_{iv}^{pq}(\bar{g}_i^\top g_v)(f_i^\top\bar{f}_v)
    +w_{iv}^{qp}(g_i^\top\bar{g}_v)(\bar{f}_i^\top f_v)
    +w_{iv}^{qq}(g_i^\top g_v)(\bar{f}_i^\top\bar{f}_v).
    \end{aligned}
    \]

    \STATE Total similarity:
    \[
    S_{iv} = \tau^{-2}(S^{F}_{iv} + S^{G}_{iv}) = 2\tau^{-2}S^{F}_{iv}
    \]

    \STATE Per-sample influence:
    \[
    \mathrm{Inf}(i) = \max_{v} S_{iv}
    \]

    \STATE Append $\mathrm{Inf}(i)$ to $\mathcal{I}$
\ENDFOR

\end{algorithmic}
\end{algorithm}

%% file: appendix/real_examples.tex
\clearpage
\section{Real Examples Sorted by InfCLIP score}\label{sec:Real_examples}
In this section, we present qualitative examples from the CC3M dataset ranked by InfCLIP score, computed using ImageNet as the target dataset.

\begin{figure}[h]
    \centering
    \includegraphics[width=0.95\linewidth]{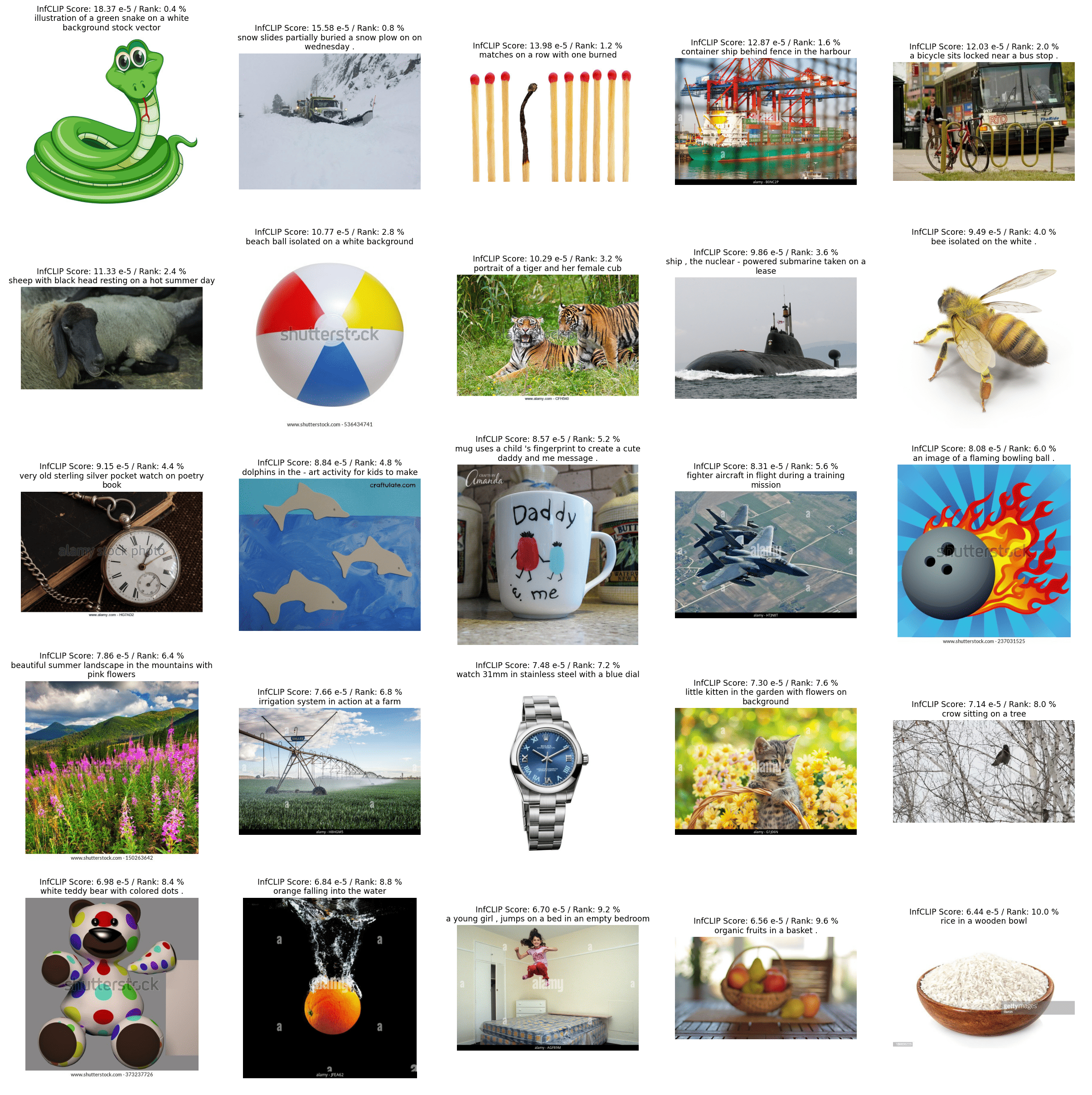}
    \caption{Examples with \textbf{high InfCLIP} scores on CC3M (ImageNet as the target).
    Samples are ordered from the highest-ranked (top-left) to lower-ranked within this subset (bottom-right).}
    \label{fig:Real_examples_high}
\end{figure}

\begin{figure}[h]
    \centering
    \includegraphics[width=0.95\linewidth]{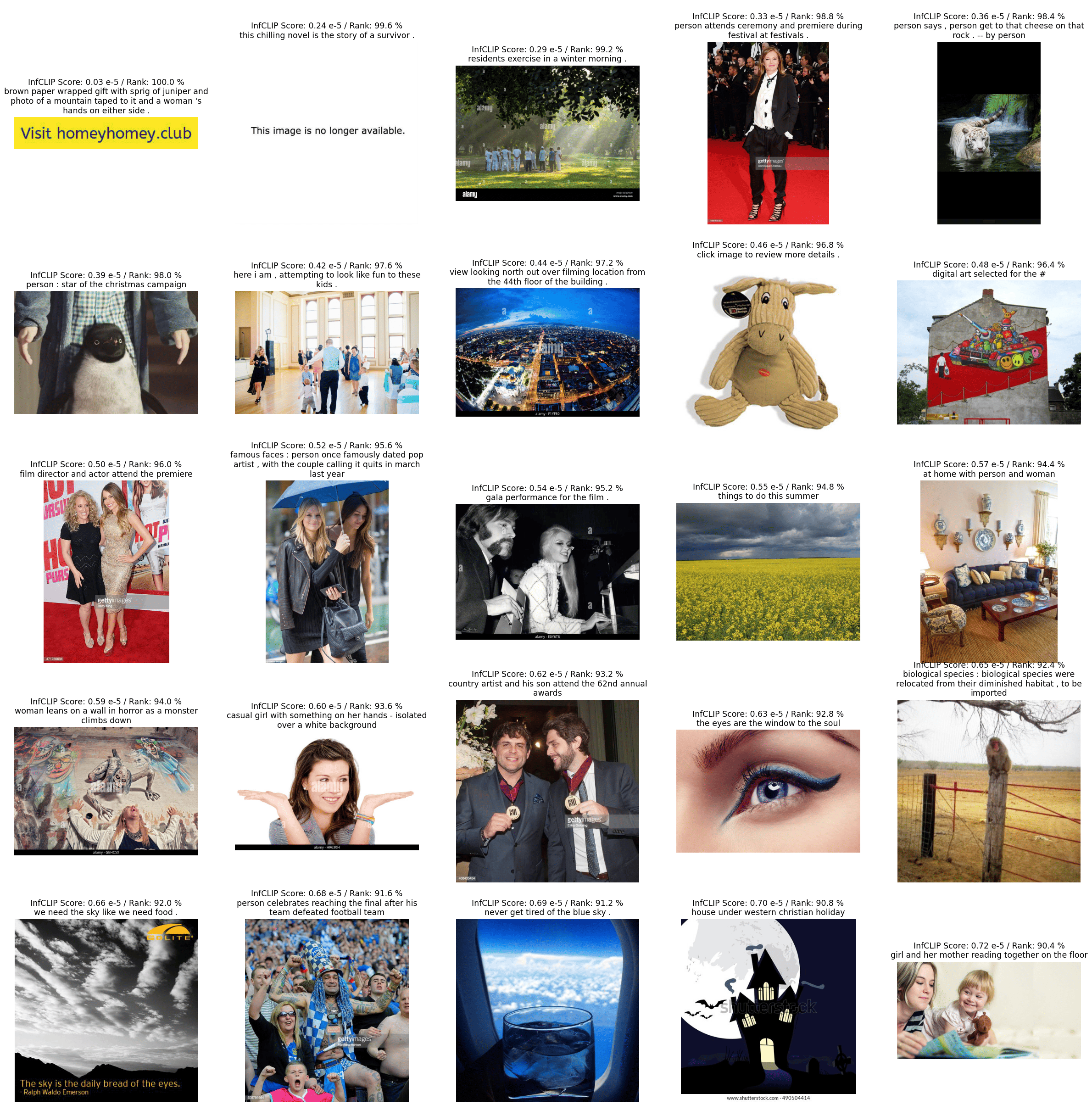}
    \caption{Examples with \textbf{low InfCLIP} scores on CC3M (ImageNet as the target).
    Samples are ordered from the lowest-ranked (top-left) to higher-ranked within this subset (bottom-right).}
    \label{fig:Real_examples_low}
\end{figure}

Figure~\ref{fig:Real_examples_high} shows examples with high InfCLIP scores.
These samples typically exhibit (i) detailed captions that accurately describe the visual content and (ii) strong relevance to ImageNet concepts (e.g., animals, vehicles, food, and common objects).
In contrast, low-scoring samples in Figure~\ref{fig:Real_examples_low} are often noisy or weakly informative.
They frequently contain mismatched image-text pairs, overly generic captions, or captions that mention objects without correctly reflecting the image content, and they tend to be less relevant to ImageNet categories.

\clearpage

%% file: appendix/full_results.tex
\clearpage
\section{Full Results}\label{app:full_result}
In this section, we present the overall results of the subset selection experiment described in Section~\ref{sec:SubsetSelection}.
For each subset ratio, we report the mean zero-shot accuracy and standard deviation across three seeds on the evaluation datasets of models trained on subsets selected by different methods.

\begin{table*}[h]
\centering
\caption{Task-wise performance of CLIP models trained on CC3M subsets at different subset ratios. COCO retrieval is evaluated using MeanRecall. Results are reported as mean $\pm$ standard deviation across three seeds.}
\resizebox{0.98\textwidth}{!}{%
\begin{tabular}{lccccccc}
\toprule
\textbf{Subset ratio 0.05} & $\mathrm{InfCLIP}$ & $\mathrm{CLIP}$ & $\mathrm{SCLIP{+}NormSim}$ & $\mathrm{HYPE{+}NormSim}$ & $\mathrm{CLIPCOV}$ & $\mathrm{ImageFilter}$ & $\mathrm{Random}$ \\
\midrule
imagenet1k & 6.41 $\pm$ 0.05 & 2.04 $\pm$ 0.04 & 5.22 $\pm$ 0.01 & 4.26 $\pm$ 0.12 & 4.76 $\pm$ 0.24 & 4.00 $\pm$ 0.09 & 1.58 $\pm$ 0.05 \\
imagenet-a & 1.73 $\pm$ 0.13 & 1.33 $\pm$ 0.14 & 1.35 $\pm$ 0.04 & 1.65 $\pm$ 0.12 & 1.43 $\pm$ 0.46 & 1.38 $\pm$ 0.13 & 0.99 $\pm$ 0.11 \\
imagenet-r & 7.67 $\pm$ 0.16 & 4.59 $\pm$ 0.08 & 4.82 $\pm$ 0.04 & 4.45 $\pm$ 0.04 & 6.52 $\pm$ 0.38 & 3.91 $\pm$ 0.07 & 2.81 $\pm$ 0.11 \\
imagenetv2 & 5.45 $\pm$ 0.06 & 1.75 $\pm$ 0.11 & 4.36 $\pm$ 0.04 & 3.77 $\pm$ 0.09 & 3.73 $\pm$ 0.50 & 3.38 $\pm$ 0.09 & 1.62 $\pm$ 0.06 \\
imagenet\_sketch & 1.51 $\pm$ 0.10 & 0.76 $\pm$ 0.10 & 0.76 $\pm$ 0.02 & 0.65 $\pm$ 0.03 & 1.53 $\pm$ 0.26 & 0.54 $\pm$ 0.03 & 0.32 $\pm$ 0.02 \\
imagenet-o & 9.71 $\pm$ 0.32 & 6.33 $\pm$ 0.38 & 8.86 $\pm$ 0.26 & 7.82 $\pm$ 0.03 & 9.85 $\pm$ 1.03 & 7.02 $\pm$ 0.33 & 5.28 $\pm$ 0.44 \\
objectnet & 2.95 $\pm$ 0.04 & 2.57 $\pm$ 0.21 & 2.09 $\pm$ 0.06 & 2.18 $\pm$ 0.04 & 3.05 $\pm$ 0.10 & 1.86 $\pm$ 0.06 & 1.58 $\pm$ 0.09 \\
vtab/cifar100 & 4.50 $\pm$ 0.89 & 2.98 $\pm$ 0.39 & 4.96 $\pm$ 0.83 & 3.71 $\pm$ 0.60 & 4.79 $\pm$ 0.77 & 3.66 $\pm$ 0.59 & 3.30 $\pm$ 0.24 \\
vtab/caltech101 & 18.72 $\pm$ 0.37 & 13.02 $\pm$ 0.91 & 15.59 $\pm$ 0.75 & 15.31 $\pm$ 0.45 & 16.71 $\pm$ 0.55 & 11.78 $\pm$ 0.58 & 8.03 $\pm$ 0.40 \\
vtab/dtd & 4.32 $\pm$ 0.65 & 3.98 $\pm$ 0.64 & 4.91 $\pm$ 0.75 & 4.18 $\pm$ 0.34 & 5.40 $\pm$ 0.72 & 4.92 $\pm$ 0.72 & 2.83 $\pm$ 0.40 \\
vtab/flowers & 7.27 $\pm$ 0.29 & 2.93 $\pm$ 0.52 & 3.81 $\pm$ 0.12 & 4.79 $\pm$ 0.60 & 3.24 $\pm$ 1.82 & 4.14 $\pm$ 0.63 & 3.05 $\pm$ 0.58 \\
vtab/pets & 8.99 $\pm$ 0.44 & 5.08 $\pm$ 0.65 & 6.99 $\pm$ 0.22 & 5.69 $\pm$ 0.29 & 5.20 $\pm$ 1.60 & 5.04 $\pm$ 0.36 & 2.74 $\pm$ 0.25 \\
sun397 & 4.82 $\pm$ 0.16 & 4.20 $\pm$ 0.41 & 6.16 $\pm$ 0.15 & 4.68 $\pm$ 0.14 & 5.02 $\pm$ 1.16 & 3.61 $\pm$ 0.18 & 4.02 $\pm$ 0.27 \\
vtab/svhn & 11.05 $\pm$ 2.39 & 14.45 $\pm$ 0.52 & 7.12 $\pm$ 0.63 & 7.99 $\pm$ 1.61 & 6.80 $\pm$ 2.28 & 12.59 $\pm$ 2.63 & 7.57 $\pm$ 3.96 \\
COCO Retrieval (MeanRecall) & 9.50 $\pm$ 0.22 & 7.59 $\pm$ 0.08 & 8.85 $\pm$ 0.06 & 6.90 $\pm$ 0.32 & 3.96 $\pm$ 0.20 & 6.22 $\pm$ 0.47 & 5.25 $\pm$ 0.44 \\
\bottomrule
\end{tabular}}
\end{table*}

\begin{table*}[h]
\centering
\resizebox{0.98\textwidth}{!}{%
\begin{tabular}{lccccccc}
\toprule
\textbf{Subset ratio 0.1} & $\mathrm{InfCLIP}$ & $\mathrm{CLIP}$ & $\mathrm{SCLIP{+}NormSim}$ & $\mathrm{HYPE{+}NormSim}$ & $\mathrm{CLIPCOV}$ & $\mathrm{ImageFilter}$ & $\mathrm{Random}$ \\
\midrule
imagenet1k & 9.57 $\pm$ 0.05 & 4.81 $\pm$ 0.05 & 7.56 $\pm$ 0.04 & 6.33 $\pm$ 0.02 & 8.53 $\pm$ 0.05 & 5.94 $\pm$ 0.04 & 3.13 $\pm$ 0.05 \\
imagenet-a & 1.62 $\pm$ 0.04 & 1.63 $\pm$ 0.06 & 2.15 $\pm$ 0.03 & 1.70 $\pm$ 0.05 & 2.19 $\pm$ 0.25 & 1.56 $\pm$ 0.05 & 1.45 $\pm$ 0.16 \\
imagenet-r & 10.97 $\pm$ 0.08 & 8.41 $\pm$ 0.03 & 6.67 $\pm$ 0.12 & 6.30 $\pm$ 0.08 & 10.42 $\pm$ 0.18 & 5.21 $\pm$ 0.08 & 3.94 $\pm$ 0.10 \\
imagenetv2 & 7.98 $\pm$ 0.07 & 4.18 $\pm$ 0.09 & 6.62 $\pm$ 0.06 & 5.46 $\pm$ 0.09 & 7.28 $\pm$ 0.28 & 5.03 $\pm$ 0.09 & 2.52 $\pm$ 0.08 \\
imagenet\_sketch & 2.58 $\pm$ 0.03 & 2.05 $\pm$ 0.12 & 1.17 $\pm$ 0.05 & 1.31 $\pm$ 0.04 & 2.88 $\pm$ 0.02 & 0.96 $\pm$ 0.02 & 0.89 $\pm$ 0.01 \\
imagenet-o & 15.42 $\pm$ 0.25 & 10.22 $\pm$ 0.21 & 10.69 $\pm$ 0.37 & 10.46 $\pm$ 0.19 & 14.69 $\pm$ 1.08 & 8.76 $\pm$ 0.24 & 7.00 $\pm$ 0.38 \\
objectnet & 4.37 $\pm$ 0.17 & 3.34 $\pm$ 0.12 & 3.24 $\pm$ 0.08 & 2.66 $\pm$ 0.14 & 5.13 $\pm$ 0.30 & 2.86 $\pm$ 0.11 & 2.35 $\pm$ 0.09 \\
vtab/cifar100 & 7.71 $\pm$ 0.52 & 5.47 $\pm$ 0.67 & 4.93 $\pm$ 0.76 & 5.35 $\pm$ 0.48 & 6.50 $\pm$ 1.06 & 4.08 $\pm$ 0.48 & 4.02 $\pm$ 0.46 \\
vtab/caltech101 & 26.33 $\pm$ 0.72 & 21.83 $\pm$ 0.29 & 21.30 $\pm$ 0.76 & 20.07 $\pm$ 0.59 & 28.52 $\pm$ 2.13 & 16.19 $\pm$ 0.46 & 14.30 $\pm$ 0.32 \\
vtab/dtd & 5.66 $\pm$ 0.16 & 6.44 $\pm$ 0.36 & 6.36 $\pm$ 0.24 & 5.78 $\pm$ 0.18 & 7.54 $\pm$ 0.76 & 4.47 $\pm$ 0.21 & 4.40 $\pm$ 0.89 \\
vtab/flowers & 10.26 $\pm$ 0.34 & 6.02 $\pm$ 0.13 & 8.01 $\pm$ 0.23 & 6.29 $\pm$ 0.47 & 5.87 $\pm$ 1.08 & 6.33 $\pm$ 0.34 & 4.77 $\pm$ 0.66 \\
vtab/pets & 11.38 $\pm$ 0.16 & 7.41 $\pm$ 0.28 & 7.35 $\pm$ 0.23 & 8.96 $\pm$ 0.38 & 10.14 $\pm$ 0.83 & 6.56 $\pm$ 0.26 & 5.21 $\pm$ 0.28 \\
sun397 & 8.72 $\pm$ 0.13 & 9.63 $\pm$ 0.18 & 11.06 $\pm$ 0.20 & 8.45 $\pm$ 0.12 & 10.06 $\pm$ 0.77 & 6.71 $\pm$ 0.12 & 5.93 $\pm$ 0.24 \\
vtab/svhn & 11.21 $\pm$ 3.71 & 9.32 $\pm$ 5.34 & 9.07 $\pm$ 2.43 & 7.17 $\pm$ 3.99 & 11.48 $\pm$ 0.24 & 8.11 $\pm$ 3.21 & 7.46 $\pm$ 3.43 \\
COCO Retrieval (MeanRecall) & 13.90 $\pm$ 0.18 & 12.04 $\pm$ 0.54 & 12.69 $\pm$ 0.25 & 11.51 $\pm$ 0.30 & 10.74 $\pm$ 0.02 & 8.37 $\pm$ 0.26 & 8.05 $\pm$ 0.43 \\
\bottomrule
\end{tabular}}
\end{table*}

\begin{table*}[h]
\centering
\resizebox{0.98\textwidth}{!}{%
\begin{tabular}{lccccccc}
\toprule
\textbf{Subset ratio 0.2} & $\mathrm{InfCLIP}$ & $\mathrm{CLIP}$ & $\mathrm{SCLIP{+}NormSim}$ & $\mathrm{HYPE{+}NormSim}$ & $\mathrm{CLIPCOV}$ & $\mathrm{ImageFilter}$ & $\mathrm{Random}$ \\
\midrule
imagenet1k & 14.38 $\pm$ 0.08 & 10.66 $\pm$ 0.04 & 12.34 $\pm$ 0.09 & 10.20 $\pm$ 0.02 & 14.07 $\pm$ 0.15 & 9.43 $\pm$ 0.06 & 6.51 $\pm$ 0.22 \\
imagenet-a & 2.81 $\pm$ 0.12 & 2.24 $\pm$ 0.14 & 1.96 $\pm$ 0.16 & 2.10 $\pm$ 0.10 & 2.40 $\pm$ 0.03 & 2.20 $\pm$ 0.11 & 1.78 $\pm$ 0.10 \\
imagenet-r & 18.00 $\pm$ 0.14 & 15.79 $\pm$ 0.13 & 11.59 $\pm$ 0.03 & 10.87 $\pm$ 0.11 & 17.35 $\pm$ 0.21 & 8.06 $\pm$ 0.08 & 8.17 $\pm$ 0.13 \\
imagenetv2 & 12.16 $\pm$ 0.02 & 8.75 $\pm$ 0.28 & 10.34 $\pm$ 0.19 & 8.72 $\pm$ 0.23 & 11.52 $\pm$ 0.11 & 7.76 $\pm$ 0.18 & 5.18 $\pm$ 0.25 \\
imagenet\_sketch & 6.52 $\pm$ 0.04 & 4.82 $\pm$ 0.06 & 3.34 $\pm$ 0.06 & 2.87 $\pm$ 0.06 & 6.67 $\pm$ 0.19 & 1.70 $\pm$ 0.03 & 2.36 $\pm$ 0.03 \\
imagenet-o & 21.42 $\pm$ 0.83 & 17.61 $\pm$ 0.38 & 18.16 $\pm$ 0.51 & 15.62 $\pm$ 0.24 & 20.70 $\pm$ 0.39 & 11.80 $\pm$ 0.29 & 9.99 $\pm$ 0.33 \\
objectnet & 6.82 $\pm$ 0.08 & 7.19 $\pm$ 0.26 & 5.55 $\pm$ 0.19 & 5.03 $\pm$ 0.20 & 8.17 $\pm$ 0.22 & 4.19 $\pm$ 0.14 & 4.00 $\pm$ 0.26 \\
vtab/cifar100 & 7.71 $\pm$ 0.44 & 10.87 $\pm$ 0.54 & 9.73 $\pm$ 0.67 & 8.80 $\pm$ 0.19 & 9.74 $\pm$ 1.33 & 6.02 $\pm$ 0.38 & 5.08 $\pm$ 0.53 \\
vtab/caltech101 & 39.26 $\pm$ 0.70 & 33.63 $\pm$ 1.34 & 32.67 $\pm$ 0.37 & 29.60 $\pm$ 0.78 & 40.26 $\pm$ 0.56 & 23.72 $\pm$ 0.37 & 22.88 $\pm$ 0.08 \\
vtab/dtd & 10.80 $\pm$ 0.52 & 13.32 $\pm$ 0.30 & 10.48 $\pm$ 0.39 & 8.52 $\pm$ 0.38 & 11.56 $\pm$ 0.07 & 5.64 $\pm$ 0.23 & 7.23 $\pm$ 0.88 \\
vtab/flowers & 11.28 $\pm$ 0.50 & 8.08 $\pm$ 0.39 & 10.52 $\pm$ 0.23 & 10.05 $\pm$ 0.63 & 9.63 $\pm$ 0.13 & 9.13 $\pm$ 0.36 & 7.07 $\pm$ 0.23 \\
vtab/pets & 13.54 $\pm$ 0.13 & 10.69 $\pm$ 0.32 & 11.89 $\pm$ 0.33 & 11.73 $\pm$ 0.23 & 11.00 $\pm$ 0.48 & 8.76 $\pm$ 0.26 & 6.37 $\pm$ 0.53 \\
sun397 & 20.19 $\pm$ 0.16 & 18.14 $\pm$ 0.64 & 20.62 $\pm$ 0.24 & 16.17 $\pm$ 0.14 & 20.93 $\pm$ 0.40 & 12.86 $\pm$ 0.14 & 12.53 $\pm$ 0.12 \\
vtab/svhn & 6.74 $\pm$ 3.12 & 10.01 $\pm$ 0.71 & 8.46 $\pm$ 2.60 & 9.72 $\pm$ 1.19 & 11.44 $\pm$ 1.45 & 11.03 $\pm$ 1.88 & 9.28 $\pm$ 1.99 \\
COCO Retrieval (MeanRecall) & 20.72 $\pm$ 0.36 & 21.30 $\pm$ 0.31 & 20.42 $\pm$ 0.25 & 18.60 $\pm$ 0.05 & 21.53 $\pm$ 0.16 & 15.16 $\pm$ 0.56 & 13.45 $\pm$ 0.62 \\
\bottomrule
\end{tabular}}
\end{table*}

\clearpage

\begin{table*}[h]
\centering
\resizebox{0.98\textwidth}{!}{%
\begin{tabular}{lccccccc}
\toprule
\textbf{Subset ratio 0.3} & $\mathrm{InfCLIP}$ & $\mathrm{CLIP}$ & $\mathrm{SCLIP{+}NormSim}$ & $\mathrm{HYPE{+}NormSim}$ & $\mathrm{CLIPCOV}$ & $\mathrm{ImageFilter}$ & $\mathrm{Random}$ \\
\midrule
imagenet1k & 17.07 $\pm$ 0.06 & 14.01 $\pm$ 0.24 & 15.16 $\pm$ 0.10 & 13.11 $\pm$ 0.04 & 16.35 $\pm$ 0.10 & 11.33 $\pm$ 0.07 & 9.31 $\pm$ 0.17 \\
imagenet-a & 2.88 $\pm$ 0.02 & 2.45 $\pm$ 0.12 & 2.87 $\pm$ 0.12 & 2.47 $\pm$ 0.22 & 2.73 $\pm$ 0.06 & 2.41 $\pm$ 0.09 & 2.31 $\pm$ 0.07 \\
imagenet-r & 21.26 $\pm$ 0.11 & 19.27 $\pm$ 0.18 & 16.18 $\pm$ 0.03 & 14.97 $\pm$ 0.13 & 20.27 $\pm$ 0.35 & 10.22 $\pm$ 0.07 & 11.37 $\pm$ 0.01 \\
imagenetv2 & 14.33 $\pm$ 0.12 & 11.66 $\pm$ 0.21 & 12.70 $\pm$ 0.02 & 11.55 $\pm$ 0.13 & 13.64 $\pm$ 0.18 & 9.66 $\pm$ 0.12 & 8.08 $\pm$ 0.20 \\
imagenet\_sketch & 8.56 $\pm$ 0.08 & 7.13 $\pm$ 0.08 & 5.76 $\pm$ 0.00 & 4.94 $\pm$ 0.06 & 8.50 $\pm$ 0.20 & 2.13 $\pm$ 0.02 & 3.59 $\pm$ 0.09 \\
imagenet-o & 24.01 $\pm$ 0.08 & 21.59 $\pm$ 0.46 & 21.22 $\pm$ 0.12 & 20.03 $\pm$ 0.47 & 24.10 $\pm$ 0.91 & 15.82 $\pm$ 0.36 & 14.32 $\pm$ 0.67 \\
objectnet & 9.77 $\pm$ 0.00 & 9.06 $\pm$ 0.32 & 8.20 $\pm$ 0.17 & 6.68 $\pm$ 0.07 & 9.09 $\pm$ 0.25 & 5.51 $\pm$ 0.13 & 5.48 $\pm$ 0.30 \\
vtab/cifar100 & 13.24 $\pm$ 0.07 & 11.97 $\pm$ 0.39 & 11.73 $\pm$ 0.58 & 8.25 $\pm$ 0.42 & 12.50 $\pm$ 1.09 & 7.36 $\pm$ 0.37 & 8.72 $\pm$ 1.33 \\
vtab/caltech101 & 42.55 $\pm$ 0.25 & 39.99 $\pm$ 0.60 & 40.86 $\pm$ 0.06 & 35.87 $\pm$ 0.70 & 44.30 $\pm$ 0.55 & 28.66 $\pm$ 0.39 & 30.04 $\pm$ 0.83 \\
vtab/dtd & 12.80 $\pm$ 0.65 & 11.83 $\pm$ 0.22 & 11.11 $\pm$ 0.35 & 10.75 $\pm$ 0.92 & 11.89 $\pm$ 0.70 & 8.53 $\pm$ 0.46 & 9.12 $\pm$ 0.67 \\
vtab/flowers & 13.35 $\pm$ 0.02 & 13.16 $\pm$ 0.97 & 12.03 $\pm$ 0.22 & 11.45 $\pm$ 0.26 & 10.63 $\pm$ 1.20 & 8.48 $\pm$ 0.41 & 8.80 $\pm$ 0.97 \\
vtab/pets & 15.14 $\pm$ 0.43 & 13.50 $\pm$ 0.50 & 14.46 $\pm$ 0.25 & 13.00 $\pm$ 0.31 & 12.77 $\pm$ 0.23 & 10.02 $\pm$ 0.27 & 7.70 $\pm$ 0.32 \\
sun397 & 27.47 $\pm$ 0.26 & 25.41 $\pm$ 0.62 & 26.17 $\pm$ 0.27 & 21.91 $\pm$ 0.39 & 26.10 $\pm$ 0.41 & 18.30 $\pm$ 0.24 & 18.11 $\pm$ 0.08 \\
vtab/svhn & 7.90 $\pm$ 0.49 & 6.10 $\pm$ 0.11 & 11.17 $\pm$ 1.01 & 7.83 $\pm$ 4.11 & 8.40 $\pm$ 1.70 & 10.77 $\pm$ 1.58 & 10.90 $\pm$ 3.44 \\
COCO Retrieval (MeanRecall) & 26.42 $\pm$ 0.05 & 27.30 $\pm$ 0.21 & 27.32 $\pm$ 0.08 & 24.09 $\pm$ 0.69 & 26.72 $\pm$ 0.25 & 19.67 $\pm$ 0.40 & 19.49 $\pm$ 0.89 \\
\bottomrule
\end{tabular}}
\end{table*}

\begin{table*}[h]
\centering
\resizebox{0.98\textwidth}{!}{%
\begin{tabular}{lccccccc}
\toprule
\textbf{Subset ratio 0.4} & $\mathrm{InfCLIP}$ & $\mathrm{CLIP}$ & $\mathrm{SCLIP{+}NormSim}$ & $\mathrm{HYPE{+}NormSim}$ & $\mathrm{CLIPCOV}$ & $\mathrm{ImageFilter}$ & $\mathrm{Random}$ \\
\midrule
imagenet1k & 18.09 $\pm$ 0.07 & 16.67 $\pm$ 0.22 & 16.77 $\pm$ 0.11 & 14.64 $\pm$ 0.10 & 17.40 $\pm$ 0.15 & 13.29 $\pm$ 0.11 & 11.42 $\pm$ 0.19 \\
imagenet-a & 3.17 $\pm$ 0.08 & 2.90 $\pm$ 0.14 & 3.22 $\pm$ 0.12 & 2.99 $\pm$ 0.19 & 2.93 $\pm$ 0.11 & 2.87 $\pm$ 0.12 & 2.68 $\pm$ 0.14 \\
imagenet-r & 22.52 $\pm$ 0.14 & 21.07 $\pm$ 0.20 & 20.42 $\pm$ 0.10 & 17.94 $\pm$ 0.18 & 20.96 $\pm$ 0.19 & 11.33 $\pm$ 0.08 & 13.85 $\pm$ 0.04 \\
imagenetv2 & 15.10 $\pm$ 0.09 & 13.45 $\pm$ 0.21 & 14.10 $\pm$ 0.06 & 12.23 $\pm$ 0.19 & 14.77 $\pm$ 0.12 & 10.96 $\pm$ 0.11 & 9.30 $\pm$ 0.16 \\
imagenet\_sketch & 9.44 $\pm$ 0.10 & 8.68 $\pm$ 0.15 & 7.47 $\pm$ 0.01 & 6.54 $\pm$ 0.10 & 8.88 $\pm$ 0.13 & 2.98 $\pm$ 0.03 & 4.87 $\pm$ 0.09 \\
imagenet-o & 26.10 $\pm$ 0.21 & 24.21 $\pm$ 0.43 & 23.10 $\pm$ 0.27 & 22.11 $\pm$ 0.44 & 23.65 $\pm$ 0.61 & 18.75 $\pm$ 0.35 & 17.21 $\pm$ 0.63 \\
objectnet & 10.51 $\pm$ 0.14 & 10.02 $\pm$ 0.31 & 9.22 $\pm$ 0.21 & 8.10 $\pm$ 0.28 & 9.81 $\pm$ 0.24 & 5.82 $\pm$ 0.14 & 7.04 $\pm$ 0.27 \\
vtab/cifar100 & 14.80 $\pm$ 0.66 & 13.22 $\pm$ 0.40 & 14.14 $\pm$ 0.97 & 10.37 $\pm$ 0.95 & 10.42 $\pm$ 0.82 & 10.48 $\pm$ 0.67 & 8.85 $\pm$ 0.96 \\
vtab/caltech101 & 43.32 $\pm$ 0.20 & 42.26 $\pm$ 0.43 & 42.00 $\pm$ 0.53 & 39.06 $\pm$ 0.99 & 44.57 $\pm$ 0.45 & 32.38 $\pm$ 0.41 & 33.46 $\pm$ 0.71 \\
vtab/dtd & 11.54 $\pm$ 0.66 & 14.85 $\pm$ 0.48 & 11.51 $\pm$ 0.22 & 11.42 $\pm$ 1.16 & 13.17 $\pm$ 0.63 & 9.40 $\pm$ 0.48 & 9.22 $\pm$ 0.57 \\
vtab/flowers & 14.99 $\pm$ 0.53 & 14.56 $\pm$ 1.33 & 13.58 $\pm$ 0.46 & 12.81 $\pm$ 0.56 & 13.65 $\pm$ 0.81 & 11.22 $\pm$ 0.64 & 9.94 $\pm$ 0.89 \\
vtab/pets & 13.15 $\pm$ 0.41 & 13.83 $\pm$ 0.54 & 13.46 $\pm$ 0.22 & 12.95 $\pm$ 0.54 & 11.67 $\pm$ 0.33 & 10.23 $\pm$ 0.34 & 9.87 $\pm$ 0.41 \\
sun397 & 30.71 $\pm$ 0.47 & 28.71 $\pm$ 0.41 & 29.58 $\pm$ 0.32 & 24.45 $\pm$ 0.45 & 29.34 $\pm$ 0.68 & 22.12 $\pm$ 0.32 & 21.21 $\pm$ 0.14 \\
vtab/svhn & 6.40 $\pm$ 0.38 & 11.47 $\pm$ 2.15 & 8.01 $\pm$ 1.56 & 8.19 $\pm$ 2.69 & 11.38 $\pm$ 1.84 & 9.53 $\pm$ 1.72 & 6.27 $\pm$ 1.96 \\
COCO Retrieval (MeanRecall) & 28.30 $\pm$ 0.05 & 32.01 $\pm$ 0.34 & 29.71 $\pm$ 0.04 & 27.65 $\pm$ 0.57 & 28.38 $\pm$ 0.32 & 22.98 $\pm$ 0.46 & 22.14 $\pm$ 0.62 \\
\bottomrule
\end{tabular}}
\end{table*}

\begin{table*}[h]
\centering
\resizebox{0.98\textwidth}{!}{%
\begin{tabular}{lccccccc}
\toprule
\textbf{Subset ratio 0.5} & $\mathrm{InfCLIP}$ & $\mathrm{CLIP}$ & $\mathrm{SCLIP{+}NormSim}$ & $\mathrm{HYPE{+}NormSim}$ & $\mathrm{CLIPCOV}$ & $\mathrm{ImageFilter}$ & $\mathrm{Random}$ \\
\midrule
imagenet1k & 18.71 $\pm$ 0.08 & 17.29 $\pm$ 0.14 & 17.91 $\pm$ 0.12 & 15.12 $\pm$ 0.16 & 18.27 $\pm$ 0.19 & 14.67 $\pm$ 0.14 & 12.89 $\pm$ 0.20 \\
imagenet-a & 3.65 $\pm$ 0.16 & 3.16 $\pm$ 0.16 & 3.16 $\pm$ 0.09 & 3.13 $\pm$ 0.13 & 3.45 $\pm$ 0.17 & 3.09 $\pm$ 0.15 & 3.01 $\pm$ 0.23 \\
imagenet-r & 23.17 $\pm$ 0.16 & 21.91 $\pm$ 0.22 & 22.47 $\pm$ 0.18 & 19.52 $\pm$ 0.23 & 22.32 $\pm$ 0.03 & 13.54 $\pm$ 0.10 & 15.55 $\pm$ 0.08 \\
imagenetv2 & 15.94 $\pm$ 0.06 & 14.17 $\pm$ 0.18 & 14.75 $\pm$ 0.09 & 12.77 $\pm$ 0.25 & 15.18 $\pm$ 0.06 & 12.15 $\pm$ 0.09 & 10.60 $\pm$ 0.09 \\
imagenet\_sketch & 10.09 $\pm$ 0.12 & 9.81 $\pm$ 0.23 & 9.84 $\pm$ 0.03 & 7.92 $\pm$ 0.16 & 9.69 $\pm$ 0.07 & 4.12 $\pm$ 0.04 & 5.79 $\pm$ 0.05 \\
imagenet-o & 25.64 $\pm$ 0.33 & 25.02 $\pm$ 0.36 & 25.35 $\pm$ 0.45 & 21.60 $\pm$ 0.34 & 24.66 $\pm$ 0.34 & 21.10 $\pm$ 0.32 & 20.30 $\pm$ 0.55 \\
objectnet & 10.92 $\pm$ 0.29 & 10.21 $\pm$ 0.27 & 10.23 $\pm$ 0.26 & 8.57 $\pm$ 0.50 & 10.29 $\pm$ 0.22 & 7.49 $\pm$ 0.19 & 7.46 $\pm$ 0.16 \\
vtab/cifar100 & 15.28 $\pm$ 1.28 & 16.40 $\pm$ 0.46 & 16.19 $\pm$ 1.41 & 12.43 $\pm$ 1.65 & 14.86 $\pm$ 1.04 & 10.70 $\pm$ 0.82 & 10.13 $\pm$ 0.67 \\
vtab/caltech101 & 45.13 $\pm$ 0.14 & 44.84 $\pm$ 0.24 & 47.42 $\pm$ 1.14 & 40.93 $\pm$ 1.27 & 45.67 $\pm$ 0.37 & 35.83 $\pm$ 0.41 & 35.81 $\pm$ 0.55 \\
vtab/dtd & 13.52 $\pm$ 0.86 & 14.90 $\pm$ 0.69 & 13.25 $\pm$ 0.08 & 13.20 $\pm$ 1.55 & 12.31 $\pm$ 0.47 & 10.53 $\pm$ 0.51 & 11.11 $\pm$ 0.56 \\
vtab/flowers & 12.97 $\pm$ 0.90 & 13.38 $\pm$ 1.46 & 14.73 $\pm$ 0.74 & 12.16 $\pm$ 0.79 & 14.82 $\pm$ 0.08 & 10.01 $\pm$ 0.66 & 11.32 $\pm$ 0.77 \\
vtab/pets & 15.30 $\pm$ 0.51 & 12.67 $\pm$ 0.52 & 14.42 $\pm$ 0.21 & 12.37 $\pm$ 0.73 & 13.66 $\pm$ 0.52 & 10.45 $\pm$ 0.42 & 9.72 $\pm$ 0.41 \\
sun397 & 34.82 $\pm$ 0.72 & 31.58 $\pm$ 0.13 & 31.69 $\pm$ 0.37 & 24.45 $\pm$ 0.47 & 31.75 $\pm$ 0.96 & 25.13 $\pm$ 0.38 & 24.94 $\pm$ 0.23 \\
vtab/svhn & 6.42 $\pm$ 0.35 & 12.67 $\pm$ 4.51 & 10.34 $\pm$ 3.08 & 8.54 $\pm$ 1.12 & 8.03 $\pm$ 0.97 & 9.33 $\pm$ 2.00 & 10.73 $\pm$ 3.32 \\
COCO Retrieval (MeanRecall) & 30.20 $\pm$ 0.10 & 33.20 $\pm$ 0.91 & 32.06 $\pm$ 0.10 & 28.53 $\pm$ 0.36 & 29.83 $\pm$ 0.21 & 25.03 $\pm$ 0.03 & 24.48 $\pm$ 0.61 \\
\bottomrule
\end{tabular}}
\end{table*}

\begin{table*}[h]
\centering
\caption{Task-wise performance of CLIP models trained on CC12M subsets at different subset ratios. COCO retrieval is evaluated using MeanRecall. Results are reported as mean $\pm$ standard deviation across three seeds.}
\resizebox{0.98\textwidth}{!}{%
\begin{tabular}{lccccccc}
\toprule
\textbf{Subset ratio 0.05} & $\mathrm{InfCLIP}$ & $\mathrm{CLIP}$ & $\mathrm{SCLIP{+}NormSim}$ & $\mathrm{HYPE{+}NormSim}$ & $\mathrm{CLIPCOV}$ & $\mathrm{ImageFilter}$ & $\mathrm{Random}$ \\
\midrule
imagenet1k & 21.02 $\pm$ 0.14 & 4.39 $\pm$ 0.07 & 17.88 $\pm$ 0.02 & 14.91 $\pm$ 0.37 & 0.32 $\pm$ 0.01 & 12.94 $\pm$ 0.13 & 4.89 $\pm$ 0.14 \\
imagenet-a & 2.37 $\pm$ 0.16 & 1.71 $\pm$ 0.16 & 2.18 $\pm$ 0.05 & 1.97 $\pm$ 0.12 & 1.00 $\pm$ 0.28 & 1.85 $\pm$ 0.16 & 1.86 $\pm$ 0.18 \\
imagenet-r & 24.09 $\pm$ 0.42 & 11.41 $\pm$ 0.17 & 13.16 $\pm$ 0.10 & 11.63 $\pm$ 0.10 & 1.17 $\pm$ 0.06 & 9.04 $\pm$ 0.18 & 7.16 $\pm$ 0.24 \\
imagenetv2 & 17.09 $\pm$ 0.16 & 3.45 $\pm$ 0.19 & 14.75 $\pm$ 0.13 & 12.76 $\pm$ 0.25 & 0.28 $\pm$ 0.03 & 10.55 $\pm$ 0.15 & 4.18 $\pm$ 0.12 \\
imagenet\_sketch & 8.22 $\pm$ 0.46 & 2.87 $\pm$ 0.34 & 3.31 $\pm$ 0.09 & 2.78 $\pm$ 0.10 & 0.19 $\pm$ 0.03 & 1.74 $\pm$ 0.18 & 1.82 $\pm$ 0.08 \\
imagenet-o & 22.63 $\pm$ 0.65 & 10.23 $\pm$ 0.53 & 20.72 $\pm$ 0.52 & 17.47 $\pm$ 0.06 & 1.37 $\pm$ 0.12 & 14.29 $\pm$ 0.42 & 8.85 $\pm$ 0.65 \\
objectnet & 6.65 $\pm$ 0.08 & 4.92 $\pm$ 0.35 & 5.80 $\pm$ 0.16 & 4.35 $\pm$ 0.06 & 1.11 $\pm$ 0.03 & 3.41 $\pm$ 0.15 & 3.86 $\pm$ 0.20 \\
vtab/cifar100 & 7.76 $\pm$ 1.34 & 6.10 $\pm$ 0.68 & 5.61 $\pm$ 0.81 & 4.05 $\pm$ 0.57 & 1.79 $\pm$ 0.25 & 4.39 $\pm$ 0.66 & 4.26 $\pm$ 0.27 \\
vtab/caltech101 & 39.25 $\pm$ 0.67 & 23.97 $\pm$ 1.46 & 32.35 $\pm$ 1.35 & 29.55 $\pm$ 0.92 & 2.77 $\pm$ 0.29 & 24.36 $\pm$ 0.92 & 19.77 $\pm$ 0.87 \\
vtab/dtd & 5.80 $\pm$ 0.76 & 7.64 $\pm$ 1.06 & 10.14 $\pm$ 1.35 & 8.86 $\pm$ 0.62 & 2.66 $\pm$ 0.31 & 7.22 $\pm$ 0.79 & 5.28 $\pm$ 0.65 \\
vtab/flowers & 21.65 $\pm$ 0.76 & 5.95 $\pm$ 0.91 & 13.68 $\pm$ 0.39 & 12.71 $\pm$ 1.38 & 1.35 $\pm$ 0.66 & 7.78 $\pm$ 0.81 & 4.42 $\pm$ 0.73 \\
vtab/pets & 55.61 $\pm$ 2.40 & 10.18 $\pm$ 1.12 & 45.84 $\pm$ 1.25 & 41.97 $\pm$ 1.87 & 3.05 $\pm$ 0.81 & 35.61 $\pm$ 1.37 & 9.57 $\pm$ 0.76 \\
sun397 & 10.98 $\pm$ 0.31 & 7.23 $\pm$ 0.62 & 16.17 $\pm$ 0.36 & 13.40 $\pm$ 0.34 & 1.21 $\pm$ 0.24 & 8.83 $\pm$ 0.40 & 8.95 $\pm$ 0.53 \\
vtab/svhn & 8.93 $\pm$ 1.68 & 6.96 $\pm$ 0.22 & 7.93 $\pm$ 0.61 & 10.66 $\pm$ 1.87 & 9.96 $\pm$ 2.91 & 13.20 $\pm$ 2.00 & 10.45 $\pm$ 4.75 \\
COCO Retrieval (MeanRecall) & 17.13 $\pm$ 0.05 & 11.02 $\pm$ 0.11 & 17.03 $\pm$ 0.11 & 14.96 $\pm$ 0.10 & 1.05 $\pm$ 0.03 & 9.68 $\pm$ 0.19 & 8.23 $\pm$ 0.03 \\
\bottomrule
\end{tabular}}
\end{table*}

\begin{table*}[h]
\centering
\resizebox{0.98\textwidth}{!}{%
\begin{tabular}{lccccccc}
\toprule
\textbf{Subset ratio 0.1} & $\mathrm{InfCLIP}$ & $\mathrm{CLIP}$ & $\mathrm{SCLIP{+}NormSim}$ & $\mathrm{HYPE{+}NormSim}$ & $\mathrm{CLIPCOV}$ & $\mathrm{ImageFilter}$ & $\mathrm{Random}$ \\
\midrule
imagenet1k & 28.67 $\pm$ 0.15 & 10.34 $\pm$ 0.09 & 25.10 $\pm$ 0.11 & 21.15 $\pm$ 0.05 & 10.07 $\pm$ 0.06 & 19.69 $\pm$ 0.10 & 10.14 $\pm$ 0.13 \\
imagenet-a & 3.27 $\pm$ 0.07 & 2.24 $\pm$ 0.07 & 3.13 $\pm$ 0.04 & 3.06 $\pm$ 0.07 & 2.05 $\pm$ 0.20 & 2.90 $\pm$ 0.11 & 2.51 $\pm$ 0.24 \\
imagenet-r & 32.58 $\pm$ 0.21 & 21.19 $\pm$ 0.05 & 20.99 $\pm$ 0.34 & 18.45 $\pm$ 0.20 & 14.17 $\pm$ 0.21 & 13.74 $\pm$ 0.22 & 13.69 $\pm$ 0.30 \\
imagenetv2 & 23.43 $\pm$ 0.17 & 8.50 $\pm$ 0.16 & 20.12 $\pm$ 0.15 & 18.02 $\pm$ 0.25 & 8.54 $\pm$ 0.28 & 16.14 $\pm$ 0.21 & 8.71 $\pm$ 0.23 \\
imagenet\_sketch & 14.32 $\pm$ 0.14 & 7.34 $\pm$ 0.37 & 7.38 $\pm$ 0.30 & 6.14 $\pm$ 0.15 & 4.42 $\pm$ 0.03 & 4.01 $\pm$ 0.17 & 4.59 $\pm$ 0.04 \\
imagenet-o & 32.12 $\pm$ 0.45 & 18.52 $\pm$ 0.33 & 29.18 $\pm$ 0.88 & 24.55 $\pm$ 0.39 & 15.70 $\pm$ 1.00 & 20.36 $\pm$ 0.63 & 15.38 $\pm$ 0.73 \\
objectnet & 10.81 $\pm$ 0.36 & 8.67 $\pm$ 0.27 & 10.03 $\pm$ 0.20 & 7.82 $\pm$ 0.36 & 5.04 $\pm$ 0.26 & 5.94 $\pm$ 0.28 & 6.53 $\pm$ 0.23 \\
vtab/cifar100 & 13.21 $\pm$ 0.78 & 9.12 $\pm$ 0.97 & 9.07 $\pm$ 1.21 & 7.82 $\pm$ 0.61 & 4.09 $\pm$ 0.58 & 4.66 $\pm$ 0.79 & 6.01 $\pm$ 0.60 \\
vtab/caltech101 & 54.06 $\pm$ 1.29 & 41.72 $\pm$ 0.49 & 44.15 $\pm$ 1.37 & 40.48 $\pm$ 1.03 & 26.61 $\pm$ 1.73 & 33.11 $\pm$ 1.10 & 33.77 $\pm$ 0.66 \\
vtab/dtd & 11.24 $\pm$ 0.27 & 11.42 $\pm$ 0.56 & 13.49 $\pm$ 0.44 & 11.29 $\pm$ 0.31 & 6.26 $\pm$ 0.55 & 8.22 $\pm$ 0.62 & 8.97 $\pm$ 1.57 \\
vtab/flowers & 26.67 $\pm$ 0.76 & 10.09 $\pm$ 0.19 & 21.26 $\pm$ 0.53 & 18.75 $\pm$ 1.21 & 3.47 $\pm$ 0.55 & 12.49 $\pm$ 0.69 & 7.65 $\pm$ 0.92 \\
vtab/pets & 61.96 $\pm$ 0.76 & 20.95 $\pm$ 0.69 & 52.31 $\pm$ 1.44 & 46.90 $\pm$ 1.73 & 24.83 $\pm$ 1.77 & 48.66 $\pm$ 1.18 & 14.93 $\pm$ 0.70 \\
sun397 & 22.21 $\pm$ 0.29 & 16.85 $\pm$ 0.27 & 27.60 $\pm$ 0.42 & 24.20 $\pm$ 0.31 & 8.60 $\pm$ 0.57 & 15.86 $\pm$ 0.42 & 18.27 $\pm$ 0.65 \\
vtab/svhn & 8.74 $\pm$ 2.51 & 11.66 $\pm$ 5.82 & 11.65 $\pm$ 2.71 & 11.02 $\pm$ 5.33 & 9.29 $\pm$ 0.17 & 11.16 $\pm$ 3.63 & 13.07 $\pm$ 5.24 \\
COCO Retrieval (MeanRecall) & 29.19 $\pm$ 0.07 & 23.84 $\pm$ 0.02 & 31.60 $\pm$ 0.01 & 26.74 $\pm$ 0.04 & 9.59 $\pm$ 0.02 & 17.96 $\pm$ 0.05 & 17.14 $\pm$ 0.04 \\
\bottomrule
\end{tabular}}
\end{table*}

\begin{table*}[h]
\centering
\resizebox{0.98\textwidth}{!}{%
\begin{tabular}{lccccccc}
\toprule
\textbf{Subset ratio 0.2} & $\mathrm{InfCLIP}$ & $\mathrm{CLIP}$ & $\mathrm{SCLIP{+}NormSim}$ & $\mathrm{HYPE{+}NormSim}$ & $\mathrm{CLIPCOV}$ & $\mathrm{ImageFilter}$ & $\mathrm{Random}$ \\
\midrule
imagenet1k & 33.67 $\pm$ 0.16 & 18.56 $\pm$ 0.06 & 30.04 $\pm$ 0.18 & 25.25 $\pm$ 0.05 & 25.94 $\pm$ 0.23 & 25.06 $\pm$ 0.19 & 16.96 $\pm$ 0.48 \\
imagenet-a & 4.44 $\pm$ 0.17 & 3.14 $\pm$ 0.17 & 4.30 $\pm$ 0.30 & 3.92 $\pm$ 0.17 & 3.50 $\pm$ 0.04 & 3.90 $\pm$ 0.16 & 3.32 $\pm$ 0.15 \\
imagenet-r & 38.70 $\pm$ 0.27 & 29.55 $\pm$ 0.22 & 29.07 $\pm$ 0.06 & 26.45 $\pm$ 0.24 & 30.84 $\pm$ 0.32 & 18.74 $\pm$ 0.24 & 22.00 $\pm$ 0.29 \\
imagenetv2 & 27.76 $\pm$ 0.05 & 14.84 $\pm$ 0.41 & 24.73 $\pm$ 0.40 & 21.31 $\pm$ 0.50 & 21.32 $\pm$ 0.17 & 20.89 $\pm$ 0.36 & 14.24 $\pm$ 0.61 \\
imagenet\_sketch & 19.70 $\pm$ 0.11 & 12.46 $\pm$ 0.13 & 13.32 $\pm$ 0.21 & 11.05 $\pm$ 0.20 & 14.24 $\pm$ 0.35 & 6.86 $\pm$ 0.18 & 8.63 $\pm$ 0.11 \\
imagenet-o & 40.38 $\pm$ 1.37 & 29.73 $\pm$ 0.56 & 36.05 $\pm$ 0.88 & 30.86 $\pm$ 0.42 & 33.58 $\pm$ 0.55 & 27.87 $\pm$ 0.74 & 23.97 $\pm$ 0.69 \\
objectnet & 15.95 $\pm$ 0.16 & 13.08 $\pm$ 0.41 & 15.73 $\pm$ 0.45 & 12.31 $\pm$ 0.42 & 13.28 $\pm$ 0.31 & 9.69 $\pm$ 0.39 & 10.13 $\pm$ 0.59 \\
vtab/cifar100 & 16.78 $\pm$ 0.84 & 18.15 $\pm$ 0.79 & 16.85 $\pm$ 1.02 & 12.54 $\pm$ 0.24 & 11.49 $\pm$ 1.36 & 8.90 $\pm$ 0.86 & 10.22 $\pm$ 0.93 \\
vtab/caltech101 & 62.73 $\pm$ 0.97 & 50.56 $\pm$ 1.75 & 56.23 $\pm$ 0.56 & 51.40 $\pm$ 1.19 & 54.25 $\pm$ 0.66 & 43.00 $\pm$ 0.87 & 45.94 $\pm$ 0.13 \\
vtab/dtd & 14.70 $\pm$ 0.62 & 14.63 $\pm$ 0.28 & 17.20 $\pm$ 0.56 & 13.67 $\pm$ 0.54 & 13.37 $\pm$ 0.07 & 11.11 $\pm$ 0.54 & 11.33 $\pm$ 1.20 \\
vtab/flowers & 28.53 $\pm$ 1.11 & 14.27 $\pm$ 0.60 & 23.82 $\pm$ 0.46 & 21.80 $\pm$ 1.19 & 11.71 $\pm$ 0.14 & 15.88 $\pm$ 0.65 & 12.82 $\pm$ 0.37 \\
vtab/pets & 62.45 $\pm$ 0.51 & 33.28 $\pm$ 0.88 & 54.09 $\pm$ 1.31 & 49.17 $\pm$ 0.84 & 49.38 $\pm$ 1.87 & 47.32 $\pm$ 1.23 & 27.28 $\pm$ 1.98 \\
sun397 & 34.56 $\pm$ 0.25 & 27.59 $\pm$ 0.84 & 38.73 $\pm$ 0.39 & 33.03 $\pm$ 0.24 & 29.49 $\pm$ 0.49 & 28.82 $\pm$ 0.41 & 27.44 $\pm$ 0.23 \\
vtab/svhn & 12.38 $\pm$ 4.98 & 10.52 $\pm$ 0.65 & 12.80 $\pm$ 3.42 & 9.63 $\pm$ 1.03 & 9.21 $\pm$ 1.01 & 9.75 $\pm$ 2.13 & 8.98 $\pm$ 1.67 \\
COCO Retrieval (MeanRecall) & 38.37 $\pm$ 0.07 & 35.67 $\pm$ 0.03 & 41.62 $\pm$ 0.02 & 36.75 $\pm$ 0.04 & 30.55 $\pm$ 0.04 & 29.87 $\pm$ 0.13 & 27.87 $\pm$ 0.11 \\
\bottomrule
\end{tabular}}
\end{table*}

\begin{table*}[!t]
\centering
\resizebox{0.98\textwidth}{!}{%
\begin{tabular}{lccccccc}
\toprule
\textbf{Subset ratio 0.3} & $\mathrm{InfCLIP}$ & $\mathrm{CLIP}$ & $\mathrm{SCLIP{+}NormSim}$ & $\mathrm{HYPE{+}NormSim}$ & $\mathrm{CLIPCOV}$ & $\mathrm{ImageFilter}$ & $\mathrm{Random}$ \\
\midrule
imagenet1k & 34.46 $\pm$ 0.10 & 23.07 $\pm$ 0.35 & 30.99 $\pm$ 0.18 & 26.69 $\pm$ 0.08 & 29.62 $\pm$ 0.16 & 27.73 $\pm$ 0.20 & 20.54 $\pm$ 0.32 \\
imagenet-a & 4.96 $\pm$ 0.03 & 3.84 $\pm$ 0.16 & 4.77 $\pm$ 0.18 & 4.33 $\pm$ 0.33 & 4.51 $\pm$ 0.09 & 4.86 $\pm$ 0.15 & 3.93 $\pm$ 0.10 \\
imagenet-r & 39.84 $\pm$ 0.19 & 33.23 $\pm$ 0.27 & 34.44 $\pm$ 0.06 & 30.75 $\pm$ 0.23 & 35.42 $\pm$ 0.53 & 23.67 $\pm$ 0.22 & 26.28 $\pm$ 0.03 \\
imagenetv2 & 28.60 $\pm$ 0.22 & 18.84 $\pm$ 0.30 & 25.69 $\pm$ 0.04 & 22.38 $\pm$ 0.22 & 24.80 $\pm$ 0.29 & 22.87 $\pm$ 0.24 & 17.21 $\pm$ 0.36 \\
imagenet\_sketch & 20.42 $\pm$ 0.16 & 15.23 $\pm$ 0.14 & 17.00 $\pm$ 0.01 & 13.96 $\pm$ 0.13 & 17.23 $\pm$ 0.35 & 9.43 $\pm$ 0.18 & 10.93 $\pm$ 0.25 \\
imagenet-o & 40.95 $\pm$ 0.12 & 32.40 $\pm$ 0.60 & 39.50 $\pm$ 0.20 & 33.48 $\pm$ 0.69 & 38.38 $\pm$ 1.27 & 32.37 $\pm$ 0.67 & 27.25 $\pm$ 1.10 \\
objectnet & 18.39 $\pm$ 0.01 & 14.78 $\pm$ 0.45 & 17.75 $\pm$ 0.31 & 14.86 $\pm$ 0.14 & 16.48 $\pm$ 0.40 & 12.58 $\pm$ 0.32 & 12.99 $\pm$ 0.62 \\
vtab/cifar100 & 19.62 $\pm$ 0.09 & 20.78 $\pm$ 0.58 & 21.21 $\pm$ 0.91 & 15.19 $\pm$ 0.67 & 17.07 $\pm$ 1.29 & 13.72 $\pm$ 0.88 & 13.24 $\pm$ 1.75 \\
vtab/caltech101 & 64.55 $\pm$ 0.33 & 54.87 $\pm$ 0.72 & 58.87 $\pm$ 0.07 & 54.89 $\pm$ 0.94 & 59.27 $\pm$ 0.64 & 47.08 $\pm$ 0.65 & 50.47 $\pm$ 1.21 \\
vtab/dtd & 15.69 $\pm$ 0.69 & 15.71 $\pm$ 0.25 & 18.49 $\pm$ 0.50 & 15.73 $\pm$ 1.17 & 15.60 $\pm$ 0.80 & 13.39 $\pm$ 0.71 & 13.33 $\pm$ 0.86 \\
vtab/flowers & 26.88 $\pm$ 0.03 & 15.79 $\pm$ 1.02 & 24.52 $\pm$ 0.38 & 21.12 $\pm$ 0.41 & 15.20 $\pm$ 1.49 & 20.12 $\pm$ 0.77 & 13.57 $\pm$ 1.30 \\
vtab/pets & 64.08 $\pm$ 1.58 & 37.76 $\pm$ 1.20 & 52.93 $\pm$ 0.79 & 48.66 $\pm$ 1.02 & 52.46 $\pm$ 0.81 & 53.71 $\pm$ 1.10 & 33.05 $\pm$ 1.20 \\
sun397 & 39.79 $\pm$ 0.33 & 33.41 $\pm$ 0.71 & 41.39 $\pm$ 0.37 & 35.96 $\pm$ 0.56 & 37.02 $\pm$ 0.51 & 34.18 $\pm$ 0.43 & 33.18 $\pm$ 0.12 \\
vtab/svhn & 17.56 $\pm$ 0.95 & 9.11 $\pm$ 0.14 & 12.02 $\pm$ 0.95 & 11.89 $\pm$ 5.43 & 8.20 $\pm$ 1.45 & 12.74 $\pm$ 1.89 & 8.97 $\pm$ 2.47 \\
COCO Retrieval (MeanRecall) & 42.02 $\pm$ 0.18 & 40.17 $\pm$ 0.09 & 44.76 $\pm$ 0.00 & 40.35 $\pm$ 0.00 & 37.72 $\pm$ 0.00 & 36.89 $\pm$ 0.16 & 32.84 $\pm$ 0.01 \\
\bottomrule
\end{tabular}}
\end{table*}

\begin{table*}[!t]
\centering
\resizebox{0.98\textwidth}{!}{%
\begin{tabular}{lccccccc}
\toprule
\textbf{Subset ratio 0.5} & $\mathrm{InfCLIP}$ & $\mathrm{CLIP}$ & $\mathrm{SCLIP{+}NormSim}$ & $\mathrm{HYPE{+}NormSim}$ & $\mathrm{CLIPCOV}$ & $\mathrm{ImageFilter}$ & $\mathrm{Random}$ \\
\midrule
imagenet1k & 33.33 $\pm$ 0.13 & 27.66 $\pm$ 0.20 & 29.27 $\pm$ 0.17 & 26.24 $\pm$ 0.24 & 31.29 $\pm$ 0.29 & 29.45 $\pm$ 0.23 & 24.53 $\pm$ 0.33 \\
imagenet-a & 5.54 $\pm$ 0.21 & 4.69 $\pm$ 0.21 & 4.89 $\pm$ 0.12 & 4.59 $\pm$ 0.16 & 5.28 $\pm$ 0.23 & 5.72 $\pm$ 0.20 & 4.54 $\pm$ 0.31 \\
imagenet-r & 39.57 $\pm$ 0.23 & 35.15 $\pm$ 0.30 & 36.72 $\pm$ 0.26 & 35.54 $\pm$ 0.36 & 38.00 $\pm$ 0.05 & 27.80 $\pm$ 0.23 & 30.08 $\pm$ 0.14 \\
imagenetv2 & 27.98 $\pm$ 0.09 & 23.17 $\pm$ 0.26 & 24.30 $\pm$ 0.13 & 21.82 $\pm$ 0.38 & 26.30 $\pm$ 0.09 & 25.23 $\pm$ 0.19 & 20.72 $\pm$ 0.16 \\
imagenet\_sketch & 20.31 $\pm$ 0.20 & 16.94 $\pm$ 0.35 & 18.47 $\pm$ 0.05 & 16.84 $\pm$ 0.30 & 18.76 $\pm$ 0.12 & 12.46 $\pm$ 0.19 & 13.90 $\pm$ 0.11 \\
imagenet-o & 41.65 $\pm$ 0.47 & 36.65 $\pm$ 0.46 & 39.57 $\pm$ 0.61 & 34.55 $\pm$ 0.47 & 40.17 $\pm$ 0.49 & 34.24 $\pm$ 0.54 & 32.23 $\pm$ 0.76 \\
objectnet & 19.25 $\pm$ 0.45 & 17.42 $\pm$ 0.40 & 18.50 $\pm$ 0.40 & 15.61 $\pm$ 0.80 & 18.14 $\pm$ 0.34 & 14.87 $\pm$ 0.44 & 14.78 $\pm$ 0.27 \\
vtab/cifar100 & 22.88 $\pm$ 1.67 & 23.74 $\pm$ 0.58 & 24.32 $\pm$ 1.84 & 18.96 $\pm$ 2.19 & 19.63 $\pm$ 1.19 & 14.89 $\pm$ 1.41 & 17.00 $\pm$ 0.97 \\
vtab/caltech101 & 64.59 $\pm$ 0.18 & 60.45 $\pm$ 0.28 & 61.68 $\pm$ 1.29 & 58.22 $\pm$ 1.57 & 61.25 $\pm$ 0.42 & 55.76 $\pm$ 0.75 & 55.84 $\pm$ 0.74 \\
vtab/dtd & 16.70 $\pm$ 0.92 & 17.94 $\pm$ 0.72 & 19.38 $\pm$ 0.11 & 16.84 $\pm$ 1.72 & 17.68 $\pm$ 0.58 & 16.82 $\pm$ 0.79 & 14.95 $\pm$ 0.65 \\
vtab/flowers & 23.29 $\pm$ 1.40 & 17.82 $\pm$ 1.69 & 20.19 $\pm$ 0.88 & 19.44 $\pm$ 1.09 & 18.84 $\pm$ 0.09 & 19.86 $\pm$ 1.01 & 15.22 $\pm$ 0.90 \\
vtab/pets & 55.05 $\pm$ 1.61 & 46.30 $\pm$ 1.65 & 48.72 $\pm$ 0.62 & 44.04 $\pm$ 2.25 & 49.92 $\pm$ 1.65 & 49.89 $\pm$ 1.53 & 38.75 $\pm$ 1.42 \\
sun397 & 44.18 $\pm$ 0.79 & 40.25 $\pm$ 0.15 & 41.06 $\pm$ 0.41 & 35.58 $\pm$ 0.59 & 41.76 $\pm$ 1.10 & 38.83 $\pm$ 0.56 & 37.71 $\pm$ 0.31 \\
vtab/svhn & 7.37 $\pm$ 0.35 & 9.42 $\pm$ 2.92 & 9.54 $\pm$ 2.47 & 7.96 $\pm$ 0.91 & 7.11 $\pm$ 0.74 & 9.19 $\pm$ 1.95 & 15.99 $\pm$ 4.31 \\
COCO Retrieval (MeanRecall) & 44.08 $\pm$ 0.04 & 44.01 $\pm$ 0.13 & 43.97 $\pm$ 0.33 & 41.60 $\pm$ 0.12 & 41.78 $\pm$ 0.04 & 41.13 $\pm$ 0.05 & 37.85 $\pm$ 0.20 \\
\bottomrule
\end{tabular}}
\end{table*}